\documentclass[journal]{IEEEtai}

\usepackage[colorlinks,urlcolor=blue,linkcolor=blue,citecolor=blue]{hyperref}

\usepackage{color,array}

\usepackage{graphicx}

\usepackage{microtype}
\usepackage{subfigure}
\usepackage{booktabs} 

\usepackage{hyperref}

\usepackage{amsmath}
\usepackage{amssymb}
\usepackage{mathtools}
\usepackage{amsthm}

\usepackage{xcolor}
\usepackage{algorithm2e}
\usepackage{caption}
\usepackage{wrapfig}
\usepackage{cite}
\usepackage{multirow}
\usepackage{wrapfig}

\usepackage{amsmath,amsfonts,bm}

\def\eqref#1{equation~\ref{#1}}

\def\1{\bm{1}}

\DeclareMathAlphabet{\mathsfit}{\encodingdefault}{\sfdefault}{m}{sl}
\SetMathAlphabet{\mathsfit}{bold}{\encodingdefault}{\sfdefault}{bx}{n}

\DeclareMathOperator*{\argmax}{arg\,max}

\theoremstyle{plain}
\newtheorem{theorem}{Theorem}[section] 

\theoremstyle{definition}

\newtheorem{assumption}[theorem]{Assumption}
\theoremstyle{remark}
\newtheorem{remark}[theorem]{Remark}

\newcommand{\Dtrain}{\ensuremath{\mathcal{D}_{\text{train}}}}
\newcommand{\edit}[1]{{#1}}

\begin{document}

\title{An Explainable \edit{and} Actionable Mistrust Scoring Framework for Model Monitoring
}

\author{Nandita Bhaskhar \{nanbhas@stanford.edu\}, Daniel L. Rubin, and Christopher Lee-Messer
\thanks{This work was supported in part by the Wu Tsai Neuroscience Institute, Stanford University and in part by a collaboration with LVIS, Inc.}
\thanks{Nandita Bhaskhar is with the Department of Electrical Engineering, Stanford University, CA 94305 USA (\textit{Corresponding author}, e-mail: nanbhas@stanford.edu).}
\thanks{Daniel L. Rubin is with the Departments of Biomedical Data Science, Radiology,
and Medicine (Biomedical Informatics) and, by courtesy, Computer Science and Ophthalmology, Stanford University, CA 94305 USA (e-mail: dlrubin@stanford.edu).}
\thanks{Christopher Lee-Messer is with the Department of Neurology, Stanford University, CA 94305 USA (e-mail: cleemess@stanford.edu).}
}

\markboth{Bhaskhar \MakeLowercase{\textit{et al.}}: TRUST-LAPSE: An Explainable and Actionable Mistrust Scoring Framework for Model Monitoring} 
{Bhaskhar \MakeLowercase{\textit{et al.}}: TRUST-LAPSE: An Explainable and Actionable Mistrust Scoring Framework for Model Monitoring 
} 

\maketitle

\begin{abstract}
Continuous monitoring of trained ML models to determine when their predictions 
should and should not be trusted is essential for their safe deployment.
Such a framework ought to be high-performing, explainable, post-hoc and actionable.
We propose TRUST-LAPSE, a ``mistrust" scoring framework for continuous model monitoring. We assess the trustworthiness of each input sample's model prediction using a sequence of latent-space embeddings.  
Specifically, (a) our latent-space mistrust score estimates mistrust using distance metrics (Mahalanobis distance) and similarity metrics (cosine similarity) in the latent-space and (b) our sequential mistrust score determines deviations in correlations over the sequence of past input representations in a non-parametric, sliding-window based algorithm for actionable continuous monitoring. 
We evaluate TRUST-LAPSE via two downstream tasks: (1) distributionally shifted input detection\edit{,} and (2) data drift detection\edit{. We evaluate} across diverse domains-- audio \edit{and} vision using public datasets and further benchmark our approach on challenging, real-world electroencephalograms (EEG) datasets for seizure detection. 
Our latent-space mistrust scores achieve state-of-the-art results with AUROCs of 84.1 (vision), 73.9 (audio), \edit{and} 77.1 (clinical EEGs), outperforming baselines by over 10 points. We expose critical failures in popular baselines that remain insensitive to input semantic content, rendering them unfit for real-world model monitoring. We show that our sequential mistrust scores achieve high drift detection rates\edit{;} over 90\% of the streams show $< 20\%$ error for all domains.
Through extensive qualitative and quantitative evaluations, we show that our mistrust scores are more robust and provide explainability for easy adoption into practice. 
\end{abstract}

\begin{IEEEImpStatement}
ML models show impressive performance across different domains. However, in the real-world, they fail silently \& catastrophically, limiting their utility. With mistrust scores from TRUST-LAPSE, these failures can be automatically identified (model monitoring) and escalated to a human operator to mitigate. Our latent-space mistrust score outperforms standard baselines by over 10 points on existing benchmarks. We expose critical failures in top baselines with semantic content. In realistic settings, where data drifts over time, our sequential mistrust scores achieve very high drift detection rates and pinpoint when the model needs to be fine-tuned or retrained. As a prime example of practical impact, we’re currently assessing these
methods for EEG seizure detection and radiology tasks in clinical settings. This could pave the way for effective clinical model deployment and human-AI partnership.
\end{IEEEImpStatement}


\section{Introduction}
\label{introduction} 

\IEEEPARstart{M}{odern} machine learning (ML) has seen tremendous success in various tasks across multiple domains, surpassing human performance in many benchmarks \cite{Esteva17_skinCancer, Yala2019_mammographyDL, Krizhevsky12_imagenet}. However, despite their impressive accuracies on test sets even after rigorous validation and testing, black-box deep learning models fail silently and catastrophically with highly confident predictions \cite{Nguyen15_conf, Goodfellow14_conf, Guo17_calib}. 
Such silent failures have severe consequences in mission-critical domains like healthcare and autonomous driving, where errors are costly, resulting in injury and death \cite{Amodei16_safety}. 
In the wake of changing data and concept drifts in the real world, where incoming samples can come from shifted distributions or completely new distributions \cite{Liu20_energyOOD},
it is imperative that we continuously monitor models after deployment to assess when their predictions should not be trusted. 

While the concept of trust can be nuanced and context-driven \cite{Lipton16a_mythos}, a (mis)trust scoring framework for continuous model monitoring must satisfy the following desiderata-- it must be: \textbf{post-hoc} - use the trained, deployed model; \textbf{explainable} - help us understand why a prediction should or should not be trusted; and \textbf{actionable} - allow us to take an automated, concrete action for every prediction, e.g. accept or reject prediction, flag, alert a human, relabel data point\edit{s} or adaptively retrain the model. Most importantly, the framework must \textbf{perform well} to avoid algorithm aversion \cite{Hou2021_AIaversion}. It must (1) assign high ``mistrust" scores when the model is faced with large distributional shifts (out-of-distribution (OOD) data) such as noisy, corrupt data or new semantic content (semantic OOD data), and (2) assign low ``mistrust" scores on learnt distributions yet unseen data (in-distribution or InD data), allowing for generalization. 

To determine trust in model predictions, current works 
typically use (a) explainability methods (XAI) \cite{Crabbe2022_labelFreeExplainability, Riberio2016_LIME, Lundberg2017_SHAP}; which, though explainable, post-hoc and suitable for providing qualitative insights, require humans to oversee the explanations and are not directly actionable for continuous and automated model monitoring; or (b) uncertainty estimation techniques \cite{Blundell15_wtUncertainty, Hendrycks17_baseline, Guo17_calib, Malinin18_priorNW, Lakshminarayanan2017_ensembles, Gal16_MCMCdropout}; 
that are difficult to train or are computationally expensive. They are frequently non-post-hoc, requiring complicated modifications to training strategies and model architectures. Some even require exposure to labelled outlier data during training\edit{,} which are often unavailable\edit{,} while many do not scale well with input-dimensionality. And, as we \edit{will} show later in this paper (Section \ref{results_datasetStats}), most of them are insensitive to input semantic content, failing to perform well as a trust scoring framework.

\begin{figure}[ht]
\captionsetup{belowskip=-15pt}
\begin{center}
\includegraphics[width=0.95\textwidth/2]{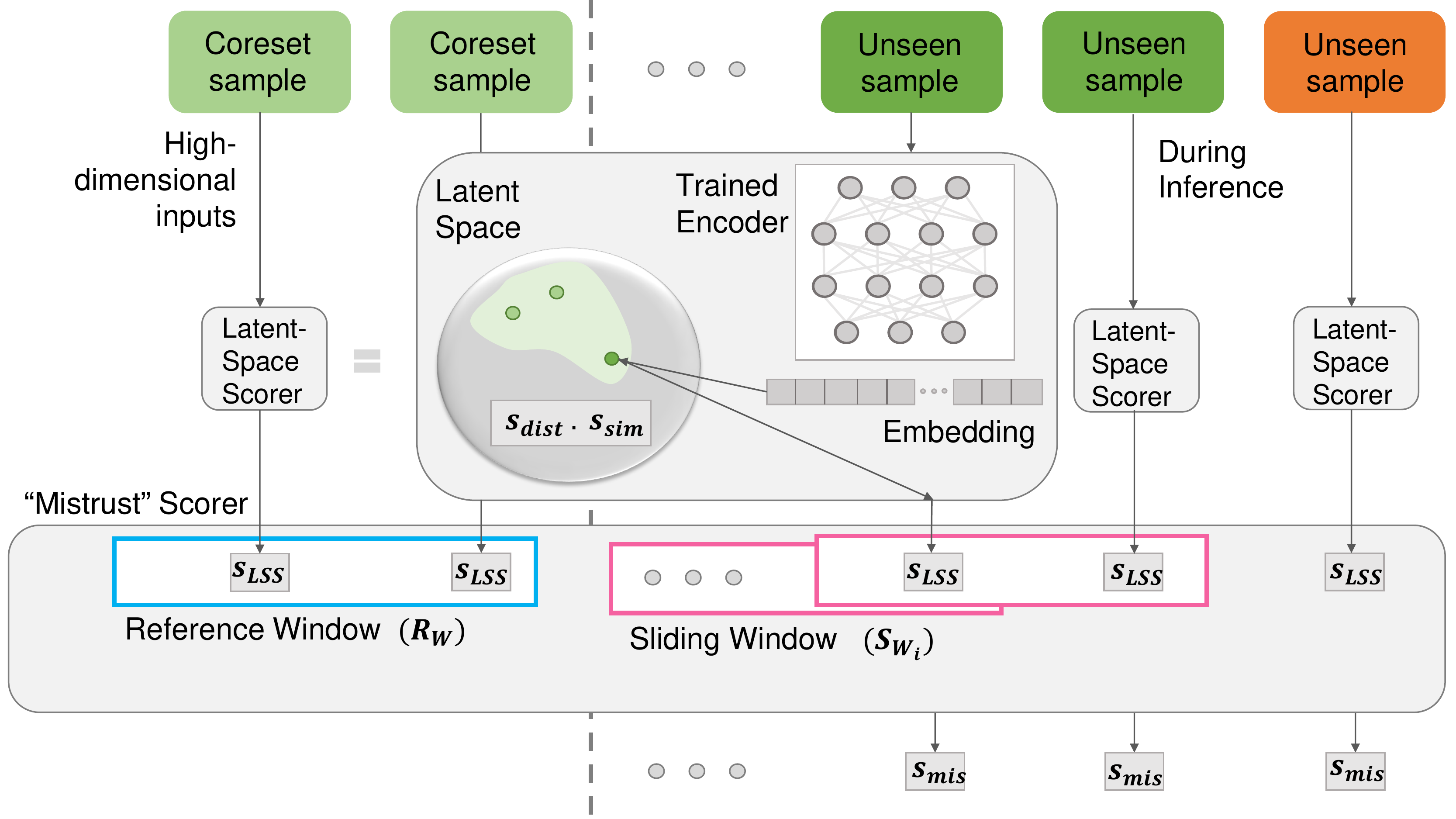}
\caption{\small TRUST-LAPSE: High-dimensional inputs (images, audio, speech \edit{and} EEG signals) are passed in sequentially during inference. Light green indicates samples from the coreset \edit{(training data subset)}. Dark green and orange represent unseen inputs, potentially trustworthy or not, respectively. Each input is projected \edit{onto} a latent-space using a well-trained encoder\edit{,} and its latent-space  mistrust score $s_{\text{LSS}}$ is extracted. The sequential mistrust scorer utilizes the sequence of latent-space mistrust scores to give the sequential mistrust $s_{\text{mis}}$.}
\label{main_figure}
\vskip -0.1in
\end{center}
\end{figure}

In this paper, we propose  TRUST-LAPSE (Fig. \ref{main_figure}), a simple framework for mistrust scoring using sequences of latent-space embeddings for continuous model monitoring. 
It is based on two key insights. First, we leverage the fact that well-trained deep learning models can act as encoders by projecting noisy, high-dimensional inputs \edit{onto} an induced hierarchical latent-space. We observe that an encoder 
with enough inductive bias will map distributionally shifted inputs differently from
other InD samples in the latent-space. 

Second, unlike standard approaches that inspect test inputs purely in isolation, we can track the trajectory of latent-space
embeddings across a set of samples to identify correlations. 
Most real-world scenarios (e.g. autonomous driving, electroencephalogram (EEG) seizure analysis, \edit{and} healthcare decision making) involve consecutive inputs to the model that are likely to be correlated sequentially e.g., an obstacle detector deployed in a self-driving car will see images correlated sequentially. A seizure detector installed in a neurology clinic will process hours of sequence-correlated EEG signals. Even low-risk models like \edit{natural} image classifiers deployed over the cloud or Netflix recommendation systems will see correlations over sequential inputs when grouped by User ID or location.
Thus, sequentially occurring samples share meaningful semantic correlations that can be leveraged for continuous model monitoring.

\vskip 0.1in

We evaluate TRUST-LAPSE via two real-world downstream tasks: (1) distributionally shifted (OOD) input detection and (2) data drift detection.
Existing benchmarks are frequently limited to highly curated image datasets and do not extend to more diverse and realistic scenarios. Furthermore, most ML systems do not evaluate their ability to detect data drifts at all, presenting a gap between models performing well on test sets and models capable of being deployed in the wild.
We perform extensive evaluations and experiments on diverse domains (audio, vision, clincal EEGs) and tasks, and show that TRUST-LAPSE is actionable, explainable and can be used with any trained model. 
Our main \textbf{contributions} are:
\begin{itemize}[leftmargin=*]
\itemsep=0em
    \item We propose  TRUST-LAPSE,  a ``mistrust" scoring framework using sequences of latent-space embeddings. To our knowledge, we are the first to use sequences of latent-space embeddings to quantify trustworthiness of model predictions for continuous model monitoring. We are also the first to use deep-learning based trust quantification techniques for EEG analyses. 
    \item We achieve state-of-the-art (SOTA) on benchmarks across diverse domains and tasks: audio speech classification (73.9 AUROC), seizure detection using clinical EEGs (77.1 AUROC) and image classification (81.4 AUROC), outperforming baselines by $\sim 10$ AUROC points.
    \item We benchmark TRUST-LAPSE and 6 other methods on their ability to flag semantically-shifted inputs as OOD while correctly identifying semantically-similar inputs (but from different datasets) as InD. 
    \item We expose critical failures in popular baselines that remain insensitive to input semantic content, unlike TRUST-LAPSE. 
    \item We achieve high detection rates when evaluating  TRUST-LAPSE on data drift detection tasks\edit{;} over 90\% of streams show $<20\%$ error on all domains. We hope to set a precedent in using \edit{data} drift detection to evaluate model trust. We believe that this is essential for characterizing and monitoring ML performance in the wild.
\end{itemize}

\section{Related Work}
\label{related_work}
\textbf{XAI methods} provide complementary, post-hoc explanations to the predictions of black-box models to induce trust. They are usually sensitivity analyses, producing systematic perturbations of the inputs to see how predictions are affected \cite{Moreno2020_uncertaintyExplainability}. They can either provide (i) feature importance explanations, that highlight top features responsible for a prediction, like LIME \cite{Riberio2016_LIME}, SHAP \cite{Lundberg2017_SHAP}, DeepLift \cite{Shrikumar2017_DeepLift}, or (ii) example importance explanations, that highlight top training examples most responsible for a prediction, like Deep K-Nearest Neighbours \cite{Papernot2018_DKNN}. \textbf{Uncertainty estimation} is a rich field with a long history.
Classical techniques like density estimation \cite{Breunig00_densityEstimation}, one-class SVMs \cite{Scholkopf99_SVM}, tree-isolation forests \cite{Liu08_isolationForest}, \edit{\cite{Jiang2018_trust}, etc., determine ``trust" using density ratios, kernels, level-set estimation and modified nearest neighbors search.}
However, they do not scale well to high dimensional datasets \cite{Rabanser19_failingLoudly}. 
Calibration is a frequentist notion of uncertainty \cite{Guo17_calib, Degroot1983_properScoring, Dawid1982_calibration}\edit{;} measured by proper scoring rules like log-loss or Brier scoring. Deep neural networks (NNs) typically use a Bayesian formalism to learn distributions over model weights \cite{Blundell15_wtUncertainty, Malinin18_priorNW, Chen19_varDirichlet, Graves2011_variationalInf, Radford1996_bayesianBook, Welling2011_bayesianLangevin} or approximate Bayesian inference such as Monte-Carlo Dropout \cite{Gal16_MCMCdropout} and Batch Normalization \cite{Teye2018_uncertainBN}. Reconstruction-based methods \cite{Zong18_deepAEanomaly, Pidhorskyi18_AEnovelty, Schlegl17_AnoGANmed, Deeke19_GANanomaly, Perera19_OCGAN} use reconstruction loss as the uncertainty score. 
Quality of uncertainty estimates are commonly evaluated via \textbf{OOD detection}, a binary classification task. 
\cite{Hendrycks17_baseline} uses maximum softmax probabilites (MSP) to detect OOD data. \cite{Liang18_ODIN} introduces a temperature parameter to the softmax equation. \cite{Lee18_mahalanobis} fits class-conditional Gaussians to intermediate activations and uses the Mahalanobis distance to identify OOD samples. Self-supervision \edit{provides} better representations that \edit{can improve} OOD detection \cite{Tack20_CSI, Winkens20_contrastiveOOD, Sehwag21_SSD}. Some \edit{works use} (semi-)supervised techniques for OOD detection with labelled outliers \cite{Hendrycks19_outlierExposure, Ruff20_semiSupervisedAD, Devries18_confEstimationBranch, Hendrycks19_selfSupervision, Mohseni20-selfSupervison, Shalev18_multipleSemanticLabelsood}. Others use
generative techniques \cite{Serra20_inputComplexityGen, Xiao20_likelihoodRegretVAE, Choi19_waic} including energy-based models \cite{Liu20_energyOOD, Du19_EBM, Grathwohl20_classifierEBM} and likelihood ratios \cite{Ren19_likelihoodRatio},
but they can be overconfident on complex inputs \cite{Nalisnick2019_highLikelihoodOOD}. 
Though many \edit{prior methods} directly optimize for good performance on OOD detection,
few \edit{can} 
serve as trust scoring frameworks for continuous model monitoring.
Our 
mistrust scores \edit{build} upon and \edit{share} similarity to \textbf{change point detection (CP) methods} in time-series data \cite{Aminikhanghahi2017_surveyChangepointDetection, Kifer04_changeDataStreams}, along with signal processing techniques such as Particle and Kalman filtering \cite{Merwe2001_particleFilter, Kalman1960_kalman}. However, traditional CP methods are limited to one (or low) dimensional data and fail in high dimensional settings \cite{Ramdas2015_temporalFailure}.
\cite{MorenoTores_datasetShift} gives a complete overview of closely related topics of distributional shift detection including covariate shift, label shift and concept drift \cite{Gama14_surveyConceptDrift}. 

\section{Preliminaries \& Problem Setup}
\label{spatio_temporal_framework-problem_setup}
Let $\mathcal{X}$ represent our high-dimensional input space, $ \mathcal{X} \subseteq \mathbb{R}^n$. Let $\mathcal{Y} = \{0, 1, 2, ..., C-1 \}$ denote the label space where $C$ is the number of classes. A black-box classification model $\mathbf{f} : \mathcal{X} \mapsto \mathcal{Y}$ is trained using a dataset $\Dtrain$ (assumed to be sampled from an underlying distribution $p^*$) such that $\mathbf{f}(x) = p(Y=y_i|x)$, where $x \in \mathcal{X}$ and $y_i \in \mathcal{Y} \; \forall \;i$. The final prediction for an unseen input $x$ during inference is given by $\hat{y} = \argmax_{y_i}{p(y_i|x)}$. 

During deployment, a steady sequence of unseen inputs $\{...,\; x_{(t-2)},\; x_{(t-1)},\; x_t,\; x_{(t+1)},\; x_{(t+2)},\; ... \}$, where $t$ denotes time, is fed to the trained model $\mathbf{f}$. For continuous model monitoring, 
our goal is to output mistrust scores $s_{\text{mis}}(x_t)$ for every input $x_t$ with threshold $\text{th}$ and a selective function $g(x_t)$  $\forall \; x_t$ such that
$$ g(x_t) :=  \begin{cases}
\hat{y_t} = \argmax_{y_i}{p(y_i|x_t)}, &\text{if} \; \; s_{\text{mis}}(x_t) \leq th \\
\text{ABSTAIN} \; \text{or} \; \text{FLAG}, &\text{else}
\end{cases} $$ 
$s_{\text{mis}}(x_t)$ is the associated mistrust in model prediction $\hat{y_t}$.

\begin{assumption}[Latent-Space Encoder]
\label{enocder_assumption}
We assume that the trained black-box  classifier $\mathbf{f}:\mathcal{X} \mapsto \mathcal{Y}$ can be decomposed as $\mathbf{f} = \mathbf{p} \circ \mathbf{h}$, where $\mathbf{h}: \mathcal{X} \mapsto \mathcal{U}$ is the derived encoder that maps high-dimensional inputs $x \in \mathcal{X} \subseteq \mathbb{R}^n$ onto a latent-space $\mathcal{U} \subseteq \mathbb{R}^d$ through its embeddings, and $\mathbf{p}: \mathcal{U} \mapsto \mathcal{Y}$ is the projection head that maps a latent-space embedding $\mathbf{u} \in \mathcal{U}$ to an output label $y = \mathbf{p}(\mathbf{u}) \in  \mathcal{Y}$. \end{assumption}

\begin{remark}
$\mathbf{h}$ is inherently learnt when $\mathbf{f}$ is trained using the InD training dataset $\Dtrain$. TRUST-LAPSE is agnostic to training strategy employed in training $\mathbf{f}$, which can be supervised, transfer-learning, self-supervised or unsupervised\edit{,} depending on the label information present in $\Dtrain$.
\end{remark}

We consider a classification setting here, though our framework can be extended to other scenarios such as regression \edit{and} segmentation.

\section{The TRUST-LAPSE Framework}
\subsection{Latent-Space Mistrust Scores}
\label{methods-spatial_score}

To extract the latent-space mistrust score of an input test sample's prediction, we rely on two assumptions: (a) a well-trained encoder maps InD samples onto a region $\mathfrak{D}$ in the latent-space but maps OOD inputs farther from this region under a distance metric, $\textit{d}$ 
, and (b) the OOD inputs that get mapped to the latent-space do not share similarity with InD inputs under a similarity metric, $\textit{sim}$. 

We first \edit{obtain} a coreset \edit{(representative subset)} from our InD training data $\Dtrain$ \edit{by random sampling,} to model the extent of the InD region $\mathfrak{D}$ within the fixed-dimensional latent-space.
$$ \text{coreset} = \{\mathbf{h}(x_i) \; | \; x_i \sim \Dtrain \}  ; \;\;\; |\Dtrain| \geq |\text{coreset}| $$

We compute the distance score $ s_\text{dist}$ and the similarity score $s_\text{sim}$ of the unseen test sample $x$ by comparing its latent-space embedding to those of samples in the coreset using a distance metric $d$ and a similarity metric $\textit{sim}$ such that:
$$ 
s_\text{dist}(x) = \min_{\mathbf{h}(x_i) \;\in\; \text{coreset}} \;\; d(\mathbf{h}(x_i), \mathbf{h}(x)) 
$$ 
$$
s_\text{sim}(x) = \max_{\mathbf{h}(x_i) \; \in \; \text{coreset}} \;\; \textit{sim}(\mathbf{h}(x_i), \mathbf{h}(x))
$$

Finally, the combined latent-space mistrust score is given by 
$$ s_\text{LSS}(x) = s_\text{dist}(x) \; . \; s_\text{sim}(x) $$

The distance score and the similarity score capture inductive biases in the latent-space differently and their product functions as an `AND' between the two. Empirically, these scores taken individually are insufficient to estimate mistrust (Section \ref{results_individual_scores}).
We note that  TRUST-LAPSE will accept any $\textit{d}$ and $\textit{sim}$ that is most suited for the target domain.
We empirically choose the distance metric $\textit{d}$ to be the Mahalanobis distance (amongst distances \edit{such as} Euclidean, Manhattan, etc.), obtained using class-conditional Gaussians \cite{Lee18_mahalanobis} without label-smoothing \cite{Winkens20_contrastiveOOD}\edit{,} where class-wise means $\mu_c$ and covariance matrices $\Sigma_c$ are estimated from the coreset. 
$$ \textit{d}(\textbf{h}(x)) = \min_{c} \; (\textbf{h}(x) - \mu_c)^T\, \Sigma_c^{-1}\, (\textbf{h}(x) - \mu_c) $$

\edit{For the} similarity metric $\textit{sim}$, we adopt the cosine similarity between the test input's representation and the coreset.
$$ \textit{sim}(\textbf{h}(x),\textbf{h}(x_i)) = \frac{\textbf{h}(x) \, . \, \textbf{h}(x_i)}{||\textbf{h}(x)|| \; ||\textbf{h}(x_i)||} ; \;\;\; \textbf{h}(x_i) \in \text{coreset}$$ 

\subsection{Sequential Mistrust Scores}
\label{methods-temporal_score}
For continuous model monitoring, where the trained model is presented with a steady sequence of high-dimensional inputs (Fig \ref{main_figure}), we begin with the assumption that there is value in evaluating samples in the context of other samples instead of viewing them in isolation. 
We use the
representational capabilities of modern encoders (Remark \ref{enocder_assumption})
and examine the sequence of latent-space mistrust scores, instead of directly modeling the incoming high-dimensional, multivariate data stream\edit{;} 
i.e., we detect when to mistrust model predictions by comparing the sequence of reduced-dimension, latent-space mistrust scores ($s_\text{LSS}$) with those of coreset samples. By ensuring that $s_\text{LSS}$ are scalar values, we reduce the problem to 
change-point (CP) detection for a one-dimensional sequence. 

We put forth an unsupervised, non-parametric, sliding-window based algorithm that builds on the work done by \cite{Kifer04_changeDataStreams} to generate our sequential mistrust scores $s_\text{mis}$. We consider a sequence of $r$ samples in the input data stream denoted by $ \{ x_1,x_2,...,x_t,...,x_r \}$. We obtain the scalar latent-space mistrust scores for the $r$ samples, $ \{ s_\text{LSS}(x_1), s_\text{LSS}(x_2),...,s_\text{LSS}(x_t),...,s_\text{LSS}(x_r) \}$ from their representations $\{ \textbf{h}(x_1),\textbf{h}(x_2),...,\textbf{h}(x_t),...,\textbf{h}(x_r) \}$ as explained in Section \ref{methods-spatial_score}. 

Given two window sizes $w_A$ and $w_B$ ($w_A,w_B \ll r$), we define a reference window $R_W$ drawn using the coreset and a sliding window from the input data stream $(S_W)_t$ to be
$$R_W = [s_\text{LSS}(x_{c_1}),...,s_\text{LSS}(x_{c_{w_A}})]$$
$$(S_W)_t = [s_\text{LSS}(x_{t}),...,s_\text{LSS}(x_{(t\,+\,w_B)})] $$ 

where $t \in \{1,...,(r-w_B)\}$ and $x_{c_i} \in \{x \;|\; \textbf{h}(x) \in \text{coreset}\}$ as visualized in Fig. \ref{main_figure} \edit{and} \ref{seq}. \edit{After} choosing an appropriate statistic or distance measure $ \mathcal{F}$ (e.g. probability odds ratio, Kolmogorov-Smirnov, Wilcoxon, Mann-Whitney, etc.), we compute the sequential mistrust score between reference window $R_W$ and the $t^{th}$ sliding window $(S_W)_t$ for the $t^{th} $ sample 
$$ s_\text{mis}(x_t) = \mathcal{F}(R_W, (S_W)_t) $$

For our empirical analysis, we chose the Mann-Whitney statistic \cite{mann_whitney} as our measure $ \mathcal{F}$. The null hypothesis $\mathcal{H}_0 $ is that the two windows (reference and sliding) come from the same distribution. We reject $\mathcal{H}_0$ in favor of the alternate hypothesis $\mathcal{H}_1 $ at a significance level of 0.05 whenever the inputs show large distribution shifts and hence the predictions should be mistrusted.
This approach (as in \cite{Kifer04_changeDataStreams}) assumes the data points are generated sequentially by some underlying probability distribution, but otherwise makes no assumptions on the nature of the generating distribution nor does it assume the samples are identically distributed.

\begin{remark}
\label{window_size}
Smaller window sizes allow changes in \edit{the} input distribution to \edit{be} reflect\edit{ed} in the mistrust scores sooner. We observed that results were relatively robust to a large range of window sizes 15-30 (vision) and 40-60 (audio, EEGs), so we \edit{chose} 25 \edit{and} 50\edit{,} respectively\edit{,} to \edit{reduce} hyperparameters. 
\end{remark}

\section{Experimental Setup}
\label{experimental_setup}

To evaluate TRUST-LAPSE, we perform experiments across multiple domains -- vision, audio and clinical EEGs. We choose the task of image classification for the visual domain and spoken word classification from audio clips for the audio domain. We further take up 
the challenging, real-world task of seizure detection from clinical EEG signals, framed as an EEG clip binary classification task for the healthcare domain. Our code \edit{is available at \url{https://github.com/nanbhas/TrustLapse}}. 

\textbf{Audio.} We use raw audio from Google Speech Commands (GSC) dataset (labels: 0-9 + words) \cite{gsc} and Free Spoken Digits Dataset (FSDD) (labels: 0-9) \cite{fsdd}. We train an M5 encoder \cite{Dai16_m5} with raw audio data from GSC 0-9 and evaluate it on (a) unseen samples from GSC 0-9 (full semantic overlap), (b) FSDD 0-9 (full semantic overlap), and (c) samples from the non-digit words from GSC (no semantic overlap).

\textbf{EEG Signals.} We train a Dense-Inception encoder \cite{Saab20_DenseInceptionEEG} on 19 channel, 200Hz sampling frequency, 60 second clips of EEG data from Stanford Health Care (SHC) (InD: patients 20-60 years \edit{of age}) and evaluate it on unseen data from SHC (full semantic overlap), Lucile Packard Children's Hospital (LPCH) (OOD partial semantic overlap: patients $<$20 years \edit{of age}) and the public Temple University Hospital seizure corpus (TUH) \cite{Shah18_TUH, Obeid16_TUH} (OOD \edit{with lower} semantic overlap: different patient demographics, \edit{and acquisition systems}).

\textbf{Vision.} We train a ResNet18 \cite{He16_resnet} encoder, trained on CIFAR10 (InD) \cite{Krizhevsky2009_cifar}, evaluated on unseen data from CIFAR10 (full semantic overlap),  CIFAR100 (partial semantic overlap) \cite{Krizhevsky2009_cifar} and SVHN (no semantic overlap) \cite{Netzer11_svhn}. We also train a LeNet encoder on MNIST (InD) \cite{Lecun1998_mnistLeNet}, and evaluate it on unseen data from MNIST (full semantic overlap), FashionMNIST (no semantic overlap) \cite{Xiao17_fashionMnist}, eMNIST (partial semantic overlap) \cite{Cohen17_emnist} and kMNIST (no semantic overlap) \cite{Tarin18_kmnist} (collectively referred to as $x$-MNIST) datasets for digit classification as is typically reported in literature \cite{Hendrycks17_baseline, Ren19_likelihoodRatio}.

Details on data pre-processing, model training, embedding extraction and sequential framework settings \edit{are given} in Appendix \ref{experimental_setup-details}. 

\vspace{5pt}
\textbf{Baselines.} We use the following baselines to compare against our framework. We do \edit{not} compare with methods that are not post-hoc, that require exposure to outliers, or are not actionable (most existing XAI methods) to match model monitoring settings.

1. MSP \cite{Hendrycks17_baseline}: Lower maximum softmax probability (MSP) $p(\hat{y}|x) = \max_k p(y=k|x)$ \edit{or} confidence indicat\edit{es} lower trust.

2. Predictive Entropy \cite{Ren19_likelihoodRatio}: Higher entropy of the predicted class distribution $- \sum_k p(y=k|x)\log p(y=k|x)$ indicates higher mistrust.

3. KL-divergence with Uniform distribution \cite{Hendrycks19_outlierExposure}: Lower KL divergence of the softmax predictions to the uniform distribution $U$, $\text{KL}(U || p(y|x))$ \edit{indicates} lower trust. 

4. ODIN \cite{Liang18_ODIN} use\edit{s} temperature-scaling and input perturbations to MSP to indicate trust\edit{.} 
We fix $T=1000, \epsilon = 1.4\text{e}^{-3}$, following their common setting, instead of tuning the parameters with outlier\edit{s}. 

5. Vanilla Mahalanobis \cite{Lee18_mahalanobis} \edit{uses} the Mahalanobis distance from the nearest class-conditional Gaussian with \textbf{\textit{shared}} covariance as the trust score. This approach directly fits in with our latent-space distance scoring method (Section \ref{methods-spatial_score}) though we use class-wise covariance matrices. 

6. Test-time Dropout \cite{Gal16_MCMCdropout} \edit{uses} Monte Carlo (test-time) dropout to estimate the prediction distribution variance to indicate uncertainty and hence, mistrust. Note that it is computationally intensive requiring $k$ forward passes of the classifier, with $k$-fold increase in runtime. We use $k=10$.

\section{TRUST-LAPSE Evaluation Framework}
Here, we describe our quantitative evaluation framework from three perspectives. In Section \ref{explainability_evaluation}, we examine TRUST-LAPSE qualitatively from the lens of explainability.

\subsection{Evaluating Latent-Space Mistrust Scores}
\label{eval_latent-space_score_framework}
For all experiments, we evaluate only using samples unseen during training.
We empirically evaluate TRUST-LAPSE's latent-space mistrust scores and all baselines in their ability to (i) assign high mistrust (flag) when the model should not be trusted, i.e., inputs with large distributional shifts the model cannot generalize to\edit{;} \edit{and} (ii) \edit{assign} low mistrust for `trustworthy inputs', i.e., inputs that the model has generalized well to.  For this task, our test sets include unseen samples drawn from the same distribution as the training data along with samples from unseen classes (new semantic content) and significantly different datasets (large distributional shifts). We evaluate performance using area under the receiver operating characteristic (AUROC), 
area under the precision-recall curve (AUPR) and the false positive rate at $80\%$ true positive rate (FPR80)\edit{, results in Section \ref{results_main}.}. Higher AUROC \edit{and} AUPR \edit{scores}  indicate better ability to differentiate between trustworthy and untrustworthy predictions \edit{and are preferred (denoted by $\uparrow$)}. Higher FPR80 \edit{scores, on the other hand,} indicate \edit{that} most predictions are not being trusted, leading to algorithm aversion. \edit{Thus, lower FPR80 scores are preferred (denoted by $\downarrow$)}. 

\subsection{Evaluating Sensitivity to Semantic Content}
\label{Semantic_sensitivity_eval_framework}
Ideally, we want methods to \textbf{\textit{flag}} new semantic content (\edit{regardless} of which dataset they belong to) since  model predictions on such inputs cannot be trusted, while at the same time being able to identify and \textbf{\textit{not flag}} semantically-similar inputs (even if they \edit{are} from different datasets) as inputs \edit{to which} the model has generalized. We also follow the setting recommended by \cite{AhmedCourville20_oodContext} to avoid dataset bias during training by holding out \edit{a} few classes from a dataset during training and using the held-out classes as new semantic content for evaluation (See Section \ref{results_datasetStats}). 

\subsection{Evaluating Sequential Mistrust Scores for Model Monitoring}
\label{Sequential_Eval_Framework}
To evaluate TRUST-LAPSE on data drift detection, we generate long sequences of test data by randomly sampling \edit{from the test sets} to emulate real-life data streams during continuous model monitoring.
For each trial, we generate a stream of test data from our test sets of length 10,000 samples with change points inserted every $k$ samples in each trial. 
Distribution changes are simulated by first randomly choosing InD or OOD (Bernoulli random variable with probability $p \in \{0.2, 0.5, 0.7\}$) and then randomly drawing $k \in \{50, 100, 200, 500, 1000, 5000 \} $ samples from the chosen distribution. This simulates various \edit{types} of OOD data.  
We evaluate performance metrics such as 
$\%$ error across trials and detection accuracy 
over $N = 1000$ trials. (See Section \ref{results_seq})

\section{Results}
\label{results} 

\vspace{-1em}
\subsection{SOTA Performance on Standard Benchmarks
}
\label{results_main}
\edit{Compared to all baselines,} TRUST-LAPSE's latent-space mistrust scores achieve the state-of-the-art (SOTA) on distributionally shifted input detection on all metrics across \edit{the} audio (spoken digit classification), clin\edit{i}cal EEG (seizure detection) and vision (image classification) \edit{domains} 
(Table \ref{table_main-results}). AUROC values for  TRUST-LAPSE scores are 73.9$\%$, 77.1$\%$, and 81.4$\%$, \edit{5.9}, \edit{12.4} and \edit{3.9} points higher than the \edit{best} baseline, for the tasks respectively. 
We \edit{perform} further \edit{analyses} in Section \ref{results_individual_scores} to examine where the performance benefits are from.

\begin{table*}[]
\captionsetup{skip=0pt}
\caption{\small Trust Scoring: Distributionally Shifted Input Detection Performance. Mean scores (standard deviations in parantheses) over 5 random runs are reported. Best scores in \textbf{bold}. TRUST-LAPSE achieves SOTA over all domains and tasks.}
\label{table_main-results}
\vskip -1in
\begin{center}
\begin{tabular}{@{}lccccccccc@{}}
\toprule
                    & \multicolumn{3}{c}{Audio}                                       & \multicolumn{3}{c}{EEG Data}                                    & \multicolumn{3}{c}{Vision}                                      \\ \cmidrule(l){2-10} 
Task (OOD Sets)     & \multicolumn{3}{c}{Speech Classification (Other spoken words)}  & \multicolumn{3}{c}{Seizure Detection (Other institutions)}      & \multicolumn{3}{c}{Image Classification (SVHN)}                 \\ \midrule
                    & AUROC $\uparrow$    & AUPR $\uparrow$     & FPR80 $\downarrow$  & AUROC $\uparrow$    & AUPR $\uparrow$     & FPR80 $\downarrow$  & AUROC $\uparrow$    & AUPR $\uparrow$     & FPR80 $\downarrow$  \\ \midrule
MSP                 & 62.6 (0.6)          & 52.7 (0.5)          & 51.5 (0.6)          & 35.8 (0.8)          & 42.1 (0.2)          & 75.4 (2.2)          & 76.0 (4.7)          & 77.0 (2.1)          & 35.8 (4.5)          \\
Predictive Entropy  & 61.5 (0.6)          & 51.5 (0.5)          & 51.5 (0.9)          & 39.3 (0.6)          & 49.5 (1.2)          & 74.2 (1.8)          & 76.1 (4.7)          & 75.2 (6.3)          & 35.7 (4.5)          \\
KL\_U               & 55.3 (0.5)          & 47.5 (0.2)          & 57.9 (0.7)          & 39.0 (0.3)          & 47.2 (0.5)          & 71.9 (1.0)          & 77.5 (5.2)          & 78.6 (2.9)          & 34.7 (6.0)          \\
ODIN                & 46.6 (0.6)          & 44.8 (1.1)          & 71.2 (9.2)          & 32.5 (1.5)          & 38.8 (2.2)          & 79.0 (3.5)          & 74.8 (6.8)          & 77.6 (3.1)          & 40.2 (9.5)          \\
Vanilla Mahalanobis & 68.0 (1.4)          & 63.6 (0.9)          & 52.0 (1.7)          & 63.3 (2.8)          & 65.1 (1.1)          & 52.5 (2.5)          & 73.8 (3.9)          & 78.2 (2.9)          & 47.7 (7.5)          \\
Test-Time Dropout   & 64.9 (0.3)          & 61.9 (0.7)          & 52.3 (0.3)          & 64.7 (0.4)          & 61.9 (0.2)          & 58.3 (1.2)          & 71.6 (5.7)          & 72.5 (4.2)          & 49.4 (6.9)          \\
TRUST-LAPSE (ours)  & \textbf{73.9 (0.6)} & \textbf{70.4 (0.8)} & \textbf{43.9 (0.5)} & \textbf{77.1 (0.9)} & \textbf{70.1 (0.4)} & \textbf{33.5 (3.9)} & \textbf{81.4 (1.9)} & \textbf{82.7 (2.4)} & \textbf{31.1 (3.4)} \\ \bottomrule
\end{tabular}
\end{center}
\vskip -0.1in
\end{table*}

\begin{table*}[t]
\captionsetup{skip=0pt}
\caption{\small \edit{E}xposing critical failures in baselines. In audio digit classification, spoken digits from different datasets (full semantic overlap) should \edit{be trusted and} result in \edit{higher} AUROC \edit{(Semantic-Split)}. Counterfactually, flagging them \edit{as untrustworthy} (\edit{D}ataset-split) should result in \edit{reduced} AURO. Mean over 5 random runs. Best scores in \textbf{bold}. Only TRUST-LAPSE achieves this with competitive performance.}
\label{table_semantic_gsc}
\begin{center}
\begin{small}
\begin{tabular}{@{}lcccccccc@{}}
\toprule
 AUROC & {Audio} &
  \multicolumn{1}{l}{MSP} &
  \multicolumn{1}{l}{\begin{tabular}[c]{@{}l@{}}Predictive\\ Entropy\end{tabular}} &
  \multicolumn{1}{l}{KL\_U} &
  \multicolumn{1}{l}{ODIN} &
  \multicolumn{1}{l}{\begin{tabular}[c]{@{}l@{}}Vanilla \\ Mahalanobis\end{tabular}} &
  \multicolumn{1}{l}{\begin{tabular}[c]{@{}l@{}}Test-Time \\ Dropout\end{tabular}} &
  \multicolumn{1}{l}{\begin{tabular}[c]{@{}l@{}} TRUST-LAPSE\\ (ours)\end{tabular}} \\ \midrule
Higher is better $\uparrow$ & Semantic-Split & 62.1 (0.6) & 61.1 (0.6) & 55.0 (0.6) & 46.6 (0.7) & 67.6 (1.6) & 64.4 (0.5) & \textbf{73.9 (0.6)} \\
Lower is better $\downarrow$ & Dataset-Split & 81.8 (0.5) & 82.1 (0.5) & 80.6 (0.6) & 67.5 (0.6) & \textbf{49.5 (0.8)} & 82.1 (0.2) & \textbf{50.6 (0.7) }\\ \bottomrule
\end{tabular}
\end{small}
\end{center}
\vskip -0.1in
\end{table*}

\subsection{Semantic Sensitivity: Exposing Critical Failures in Baselines}
\label{results_datasetStats}
We investigate if methods \textbf{\textit{flag}} \edit{inputs with }new semantic content even if \edit{they are} from the same dataset\edit{,} and \textbf{\textit{not flag}} inputs with large semantic overlaps even if \edit{they are }from different datasets. 

\begin{figure}[h]
\captionsetup{belowskip=-5pt}
\includegraphics[width=0.5\textwidth]{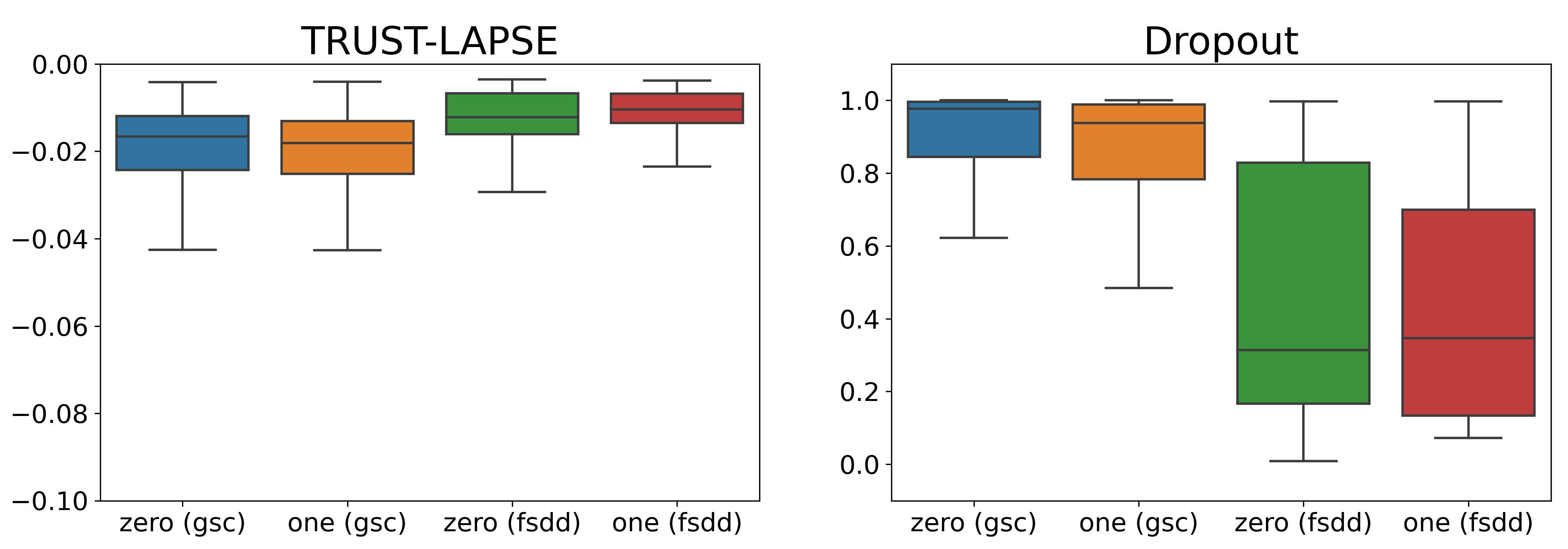}
\caption{\small TRUST-LAPSE scores inform model generalization. Audio (left) TRUST-LAPSE correctly assigns high trust to classes the model has generalized to (e.g. zero, one), \edit{regardless} of dataset (right) Test-Time Dropout assigns low trust to audio samples from FSDD dataset even though the model is capable of making correct predictions on it. }
\label{GSC_SemanticVsDataset_boxplot}
\end{figure}

\textbf{Audio.} In our audio experiments, we train our encoder only on GSC 0-9 (forming the InD classes) and find that it generalizes well to FSDD 0-9. 
The other 25 classes in GSC (Yes, No, Right, Left, Bird, etc.: GSC-Words) have never been encountered by the model \edit{and} have drastic semantic shifts in their content when compared to the InD classes 0-9\edit{.P}redictions on these classes need to be \textbf{\textit{flagged}}. 
During evaluation, we \edit{test} the model \edit{on} unseen samples from each of the classes in GSC \edit{and} FSDD 0-9 (same semantic content as InD data but different dataset) and measure which predictions are flagged as untrustworthy. We want methods to flag predictions on GSC-Words yet trust predictions on GSC 0-9 and FSDD 0-9 (Table \ref{table_semantic_gsc}, row Semantic-Split). 
For comparison, we consider the undesirable counterfactual setting where a method only trust\edit{s} predictions on GSC 0-9, flagging everything else (even though the model has generalized to predict well over FSDD 0-9).
For the counterfactual setting, we see that AUROC values for most baselines, notably Test-time Dropout and Predictive Entropy, increase significantly, whereas they drop for TRUST-LAPSE and \edit{V}anilla Mahalanobis (Table \ref{table_semantic_gsc}, row Dataset-Split). This shows popular baselines have narrow capabilities that \edit{do not respect model generalization. This is because they }assume only data belonging to InD classes \textit{and} having the \textit{\textbf{same dataset statistics}} as the training data will \edit{be} trusted, a critical failure. All other data will be flagged, leading to very high false positive rates (FPRs) and algorithm aversion in practice. In contrast, TRUST-LAPSE is able to \edit{respect model generalization, i.e.,} identify true trustworthy samples despite dataset statistics\edit{;} and flag new semantic content (Fig. \ref{GSC_SemanticVsDataset_boxplot}). 

\textbf{Vision (CIFAR10).} For our CIFAR10 experiments, we train our encoder only on 7 classes of CIFAR10, leaving out classes Airplane, Bird and Dog. 
We study the effect of semantic overlap between the 100 classes from CIFAR100, along with the 3 left out classes from CIFAR10, on the 7 InD classes (Automobile, Cat, Deer, Frog, Horse, Ship, Truck).
We analyse class-wise performance and we see that  TRUST-LAPSE's performance strongly correlates to semantic content overlap unlike other methods which seem to rely on dataset statistics. For example, we consider Class Pickup-Truck and Streetcar from CIFAR100. Though they are disjoint from the InD Classes Automobile and Truck in CIFAR10, they share a high degree of semantic overlap and should be considered InD, \edit{and} \textbf{\textit{not be flagged}} \edit{as untrustworthy}. Counterfactually requiring \edit{them to be} flagged should result in lower \edit{AUROC}  (Table \ref{table_cifar100_analysis}, top two rows)\edit{.}  \edit{P}erformance on classes that don't share any semantic overlap with the 7 InD classes should remain high (Table \ref{table_cifar100_analysis}, bottom two rows).

\textbf{Vision (MNIST).} We train two different encoders, one trained on all digits 0-9 while the other was trained on $\{0,1,4,6,7,8,9\}$ leaving \edit{out} digits 2, 3 and 5. Only certain classes from the e-MNIST data \edit{have} semantic overlap -- letters `o', `l' \edit{and} `i', `z', `y', `s'\edit{,} and `q' share structural similarity with classes 0, 1, 2, 4, 5 and 9 respectively. We show results for both encoders with (reduced AUROC) and without the above classes (\edit{Semantic,} increased AUROC) in counterfactual evaluations (Table \ref{table_semantic_mnist}). 

\textbf{Clinical EEGs.} While discussions on semantic issues on clinical EEGs may be out of scope for this paper, we do observe interesting semantic effects in this case as well. For instance, seizure types unseen by the network \edit{during training} generate higher mistrust scores \edit{(Fig. \ref{eeg_boxen})}.

\begin{table*}[h]
\captionsetup{skip=0pt}
\caption{\small CIFAR100 Pickup Truck \& Street Car share high semantic overlap with InD classes Truck \& Automobile. Counterfactually requiring them to be flagged should result in \edit{reduced} AUROC \edit{and} AUPR (top set). Airplane \& Fish do not show any semantic overlap with InD classes, and should \textbf{simultaneously} result in \edit{increased} AUROC \edit{and} AUPR (bottom set). Vanilla Mahalonobis and Test-Time Dropout achieve high performance in one set ({\color{blue}blue}) but not the other. KL\_U and TRUST-LAPSE achieve peak performance simultaneously in both sets (\textbf{bold}).  Mean over 5 random runs.}
\label{table_cifar100_analysis}
\begin{center}
\begin{small}
\begin{tabular}{@{}llccccccc@{}}
\toprule
\begin{tabular}[c]{@{}c@{}}(AUROC / \\ AUPR)\end{tabular} &
 CIFAR &
  MSP &
  \begin{tabular}[c]{@{}c@{}}Predictive\\ Entropy\end{tabular} &
  KL\_U &
  ODIN &
  \begin{tabular}[c]{@{}c@{}}Vanilla \\ Mahalanobis\end{tabular} &
  \begin{tabular}[c]{@{}c@{}}Test-Time \\ Dropout\end{tabular} &
  \begin{tabular}[c]{@{}c@{}} TRUST-LAPSE\\ (ours)\end{tabular} \\ \midrule
Lower is & Pickup Truck  & 79.4 / 29.2 & 79.1 / 27.6 & \textbf{69.8 / 18.1} & 81.5 / 18.2 & {\color{blue} 69.1 / 20.7} & 94.2 / 50.6 & \textbf{68.8 / 18.3} \\
better & Street Car    & 76.7 / 25.5 & 76.8 / 25.6 & \textbf{76.6 / 23.9} & 78.5 / 25.5 & {\color{blue} 65.6 /  18.1} & 83.8 / 33.7 & \textbf{75.7 / 25.6} \\ \midrule
Higher is & Airplane     & 89.1 / 48.8 & 89.4 / 51.9 & \textbf{91.3 / 57.1} & 89.3 / 54.9 & 65.8 / 19.2 & {\color{blue} 95.5 / 54.7} & \textbf{90.2 / 57.5} \\
better & Aquarium Fish  & 89.3 / 49.2 & 89.7 / 52.9 & \textbf{91.3 / 58.9} & 88.8 / 52.6 & 70.5 / 20.9 & {\color{blue} 95.5 / 54.2} & \textbf{91.1 / 57.7} \\ \bottomrule
\end{tabular}
\end{small}
\end{center}
\vskip -0.1in
\end{table*}

\begin{table*}[h]
\captionsetup{skip=0pt}
\caption{ \small MNIST \edit{Semantic Sensitivity}. 0to9 denotes encoder trained on 0-9 digits. M235 indicates encoder trained on digits 
\{0,1,4,6,7,8,9\}. ``Semantic" indicates that  e-MNIST classes with ambiguous semantic overlap on InD classes, i.e., \{l,i,o,..\} are removed from OOD evaluation. Mean over 5 random runs. Best scores in \textbf{bold}. \edit{While AUROC increases for the ``Semantic'' evaluation for all baselines, TRUST-LAPSE shows the highest performance in all cases.}}
\label{table_semantic_mnist}
\begin{center}
\begin{small}
\begin{tabular}{@{}lccccccc@{}}
\toprule
\multicolumn{1}{c}{\begin{tabular}[c]{@{}c@{}}MNIST \\ (AUROC $\uparrow$ / AUPR $\uparrow$)\end{tabular}} &
  MSP &
  \begin{tabular}[c]{@{}c@{}}Predictive\\ Entropy\end{tabular} &
  KL\_U &
  ODIN &
  \begin{tabular}[c]{@{}c@{}}Vanilla \\ Mahalanobis\end{tabular} &
  \begin{tabular}[c]{@{}c@{}}Test-Time \\ Dropout\end{tabular} &
  \begin{tabular}[c]{@{}c@{}} TRUST-LAPSE\\ (ours)\end{tabular} \\ \midrule
0to9          & 89.5 / 96.8 & 89.9 / 97.1 & 90.3 / 97.2 & 89.6 / 96.9 & 90.3 / 97.3 & \textbf{96.3} / 97.3 & \textbf{96.1 / 99.0}          \\
0to9 Semantic & 92.1 / 97.0 & 92.6 / 97.3 & 92.8 / 97.5 & 92.2 / 97.1 & 93.7 / 97.6 & 97.6 / 97.0 & \textbf{98.6 / 99.5} \\ \midrule
M235          & 90.7 / 98.2 & 91.0 / 98.3 & 90.9 / 98.3 & 90.6 / 98.2 & 91.9 / 98.5 & \textbf{97.6} / 98.6 & \textbf{97.5 / 99.6}          \\
M235 Semantic & 92.6 / 98.3 & 92.9 / 98.4 & 92.5 / 98.3 & 92.5 / 98.3 & 94.1 / 98.6 & 98.4 / 98.6 & \textbf{99.0 / 99.8}   \\ \bottomrule
\end{tabular}
\end{small}
\end{center}
\vskip -0.1in
\end{table*}

\subsection{Drift Detection with Sequential Mistrust Scores}
\label{results_seq}

\begin{figure}[h]
\includegraphics[width=0.45\textwidth]{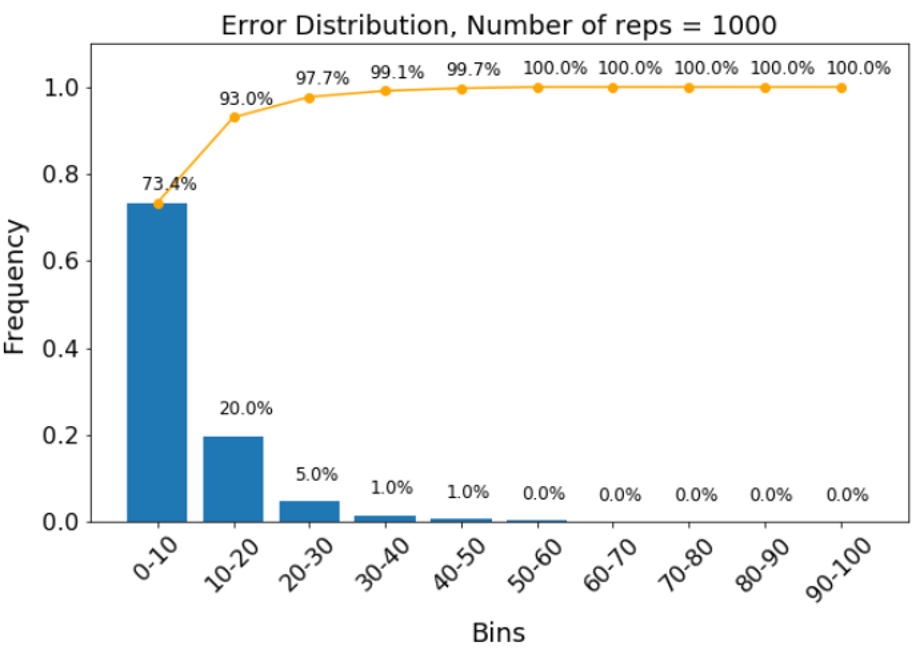}
\vskip -0.1in
\caption{\small  Drift Detection (EEG): \% Error distribution over $N=1000$ \edit{streams. $> 73\%$ of streams have $<10\%$ error; $>93\%$ have $<20\%$ error}.
}
\label{sequentialOOD_results_main}
\vskip -0.1in
\end{figure}

To evaluate the performance of TRUST-LAPSE on drift detection, we simulate 1000 data streams of 10,000 data points each for each domain, with injected change points as explained in Section \ref{Sequential_Eval_Framework}. 
We then calculate the sequential mistrust scores using the two-sided Mann-Whitney test \cite{mann_whitney} at a significance level of $0.05$ for each sample's latent-space mistrust score in the stream and run it through our sequential detector with window sizes $w_A = w_B = 25$ for vision experiments and $w_A = w_B = 50$ for audio and EEG experiments (Remark \ref{window_size}). 
We perform a 2-cluster KMeans \cite{kmeans} on the generated Mann-Whitney scores to assign \textbf{\textit{trust}} and \textbf{\textit{flag}} (mistrust) actions. 
We define the detector to have erred if it wrongly concludes that an incoming test sample's prediction is to be \textbf{\textit{flagged}} (assigned high mistrust) when it was actually supposed to be trusted 
or vice-versa. 
If the framework detects the change point within its fixed window size, we do not consider it as an error.  
For each data stream, we then add the errors of all samples to calculate detection error. We repeat this for all $N = 1000$ streams to estimate the detector's error distribution across streams.  Fig. \ref{sequentialOOD_results_main} shows the error distribution over EEG data streams. 
Over $73\%$ of the streams have less than $10\%$ error and over $93\%$ have less than $20\%$ error. 
For our audio domain, over $85\%$ of the streams have less than $10\%$ error and over $97\%$ have less than $20\%$ error.
For our vision tasks, the error distribution is much tighter and we are able to get near $99\%$ detection accuracy for over $95\%$ of the streams. 
We show plots of generated trust scores from TRUST-LAPSE and other baselines for parts of audio data streams as examples in Figs. \ref{short_plots}, \ref{audio_seq_100_main} (more plots in Appendix, Figs. \ref{sequentialOOD_results}-\ref{mnist_seq}). 

\begin{figure}[h]
\includegraphics[width=\textwidth/2]{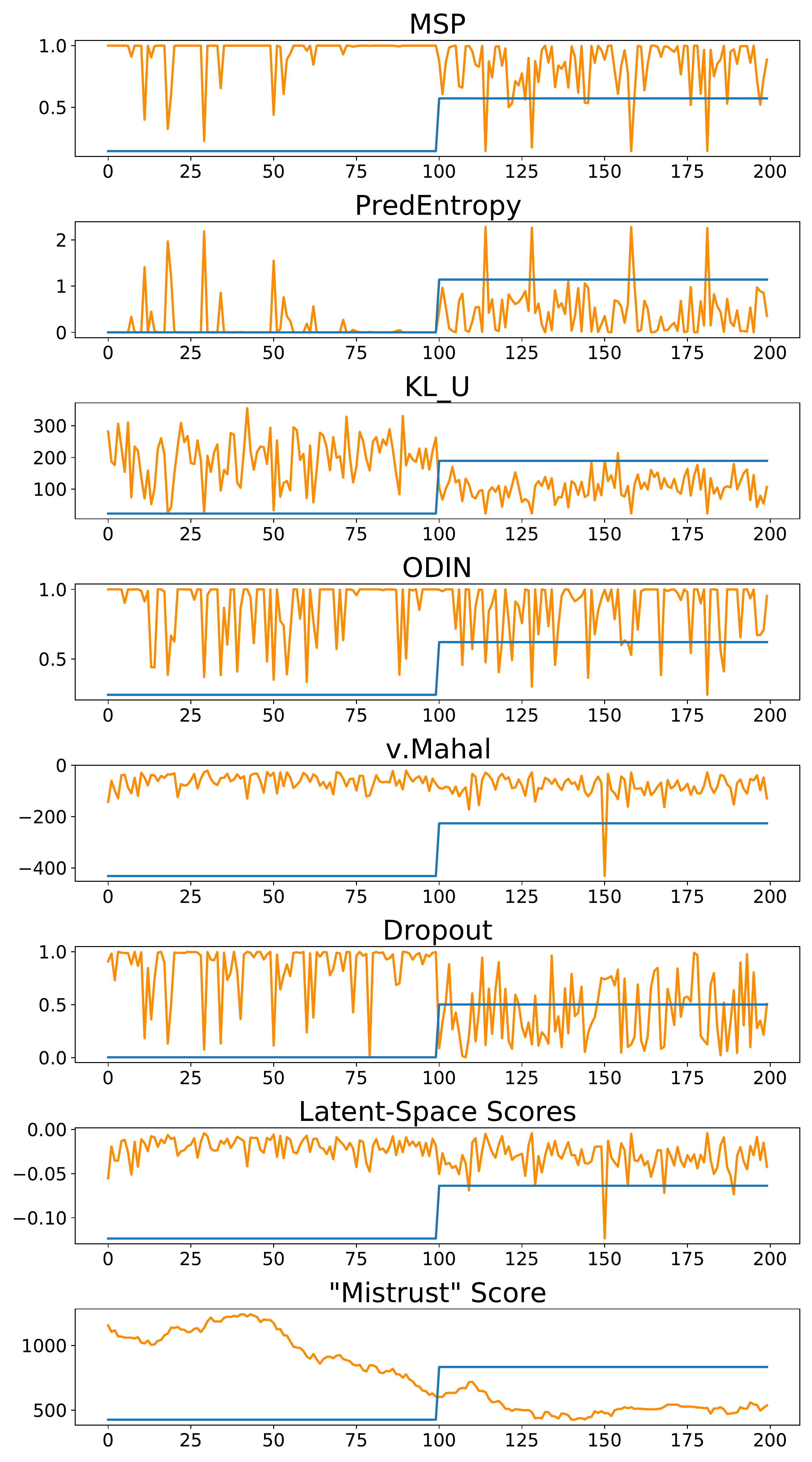}
\caption{\small Trust scores from Baselines and TRUST-LAPSE (audio).  Sequential Mistrust Score \edit{(``Mistrust'')} shows the most separability between trustworthy (blue low segment) and to be flagged (blue high segment) points. More plots in the Appendix, Figs. \ref{sequential_ablations_cifar}, \ref{sequential_ablations_mnist}.}
\label{short_plots}
\vskip -0.1in
\end{figure}

\begin{figure*}[h]
\includegraphics[width=\textwidth]{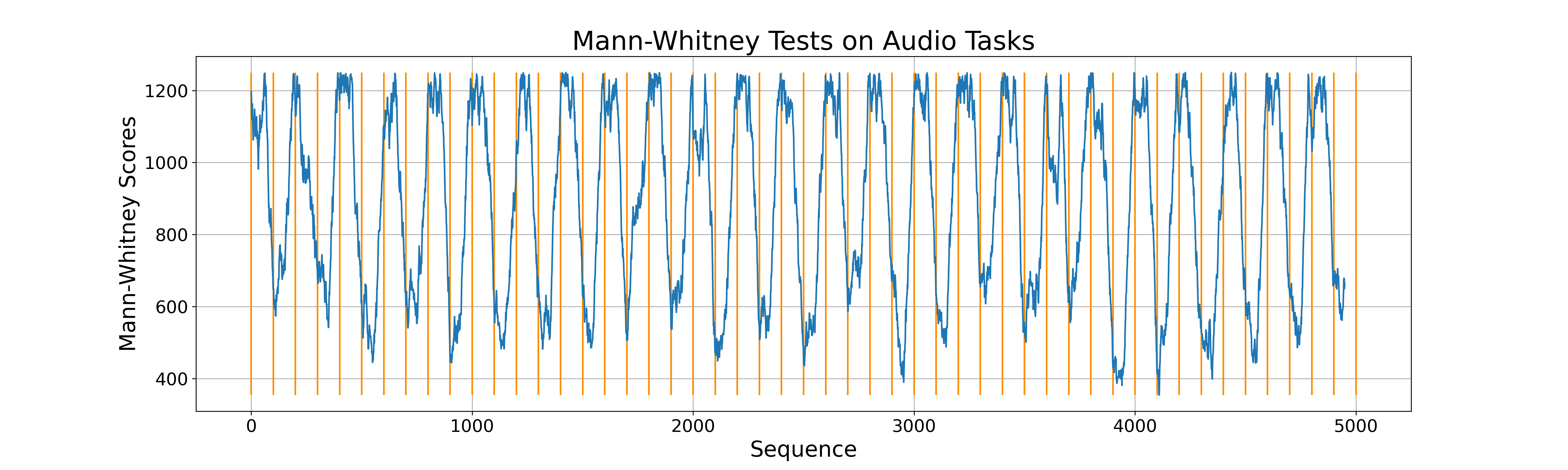}
\vskip -0.1in
\caption{\small Drift Detection Evaluation (Audio): Data stream with distribution shifts occurring every 100 steps. The orange lines indicate true change points. Blue plot represents the sequential mistrust scores (more plots in Appendix, Figs. \ref{sequentialOOD_results}-\ref{cifar_seq}).}
\label{audio_seq_100_main}

\end{figure*}

\section{Analyses and Ablations}

We perform extensive \edit{analyses with} TRUST-LAPSE to study the effects of (i) individual components in our mistrust scores (Section \ref{results_individual_scores}), (ii) coreset size (Section \ref{results_coreset_size}),  and (iii) encoder capacity (Section  \ref{results_encoder_supplementary}) on its performance.

\subsection{Importance of individual components in  TRUST-LAPSE}
\label{results_individual_scores}

We study the effect of each individual component in TRUST-LAPSE to
\edit{assess the source of its performance gains.}

\textbf{Distance Metric.} We empirically observed that Mahalanobis distance (M. Dist) outperformed Euclidean (also reported in \cite{Lee18_mahalanobis}), Manhattan, and Chebyshev distances. Unlike \cite{Winkens20_contrastiveOOD}, who use M. Dist with label-smoothing and \cite{Lee18_mahalanobis}, with shared covariance, we observed M. Dist perform\edit{s} best with class-wise covariance matrices and without label-smoothing. We found the shared covariance assumption more restrictive and did not fit different types of data. From Table \ref{table_importance_scores}, we see that it performs worse compared to our M. Dist formulation (Section \ref{methods-spatial_score}).
We also see that using only our M. Dist follows the same trend as  TRUST-LAPSE \edit{in terms of} semantic sensitivity\edit{;} i.e., it is more sensitive to semantic content than dataset statistics, but its performance is not high enough to be competitive.

\textbf{Similarity Metric.} We see that cosine similarity (C. Sim) is is sensitive to both dataset-statistics and semantic-content. But just using C. Sim can be dangerous since it can be overly sensitive to dataset statistics even if there is no semantic overlap (Table \ref{table_importance_scores}, audio). In combination with M. Dist, TRUST-LAPSE is able to reap the best of both metrics (Table \ref{table_importance_scores}).

\textbf{Latent-space mistrust ($s_\text{LSS})$ vs Sequential mistrust ($s_\text{mis}$).} Visualizing the scores for a datastream (Fig. \ref{short_plots}) shows that despite high AUROCs, $s_\text{LSS}$ \edit{(Fig. \ref{short_plots}, Latent-Space Scores)} and baselines \edit{(Fig. \ref{short_plots}, top 6 plots)} show small margins between low mistrust and high mistrust scores and are more sensitive to possible threshold changes. Adding $s_\text{mis}$ \edit{(Fig. \ref{short_plots}, ``Mistrust'' Score)} makes TRUST-LAPSE more robust and easy to interpret.

\textbf{Only sequential mistrust scores.} Conceptually, we can use our sequential windowing approach (Section \ref{methods-temporal_score}) directly on the (non-scalar) latent-space embeddings, skipping latent-space mistrust scoring (Section \ref{methods-spatial_score}) entirely. However, \cite{Ramdas2015_temporalFailure} shows that hypothesis-based tests do not scale well to the multivariate case. Hence, we do not consider it a viable option.

\begin{table}[h]
\captionsetup{skip=0pt}
\caption{\small Contributions from individual components of TRUST-LAPSE. Details on Semantic- and Dataset-Splits in Section \ref{results_datasetStats}.}
\label{table_importance_scores}
\begin{center}
\begin{tabular}{@{}llccc@{}}
\toprule
                   &        Score                                                                              & Audio & CIFAR                    & MNIST                    \\ \midrule
\multirow{4}{*}{\begin{tabular}[l]{@{}l@{}}Semantic\\ Split\\ AUROC $\uparrow$\end{tabular}} & Only distance (M. Dist \cite{Lee18_mahalanobis}) &  65.8  & 96.2                     & 95.1                     \\
                        & Only distance (Our\edit{s,} M. Dist)                       & 72.8  & 98.0                     & 96.2                     \\
                        &   Only similarity (\edit{O}urs\edit{, C. Sim})                          & 65.8  & 98.4                     & 98.9                     \\
                        &  Latent-Space Mistrust (\edit{O}urs)                    & 73.9  & 98.9                     & 99.0                     \\ \midrule
\multirow{4}{*}{\begin{tabular}[l]{@{}l@{}}Dataset\\ Split\\ AUROC $\downarrow$\end{tabular}}  & Only distance (M. Dist \cite{Lee18_mahalanobis}) &  48.0  & 69.4 & 91.9 \\
  &   Only distance (Our M. Dist)                                                                                   & 50.2  & 83.2 & 94.8 \\
   &   Only similarity (\edit{O}urs\edit{, C. Sim})                                                                                   & 81.7  & 88.0 & 97.0 \\
   &   Latent-Space Mistrust (\edit{O}urs)                                                                                   & 50.6  & 87.5 & 97.4 \\
\bottomrule
\end{tabular}
\end{center}
\vskip -0.1in
\end{table}

\subsection{Coreset size}
\label{results_coreset_size}

Our latent-space mistrust score uses cosine similarity for pairwise comparison of a test sample's latent-space embedding with those of training data. This could be computationally intensive, especially for large datasets. We reduce computation costs, memory overhead and latency by using a subset of training data (coreset) instead of the full train\edit{ing }set. The coreset is extracted simply by randomly sampling a fraction from every class in the train\edit{ing }set, unlike complicated strategies used in other works \cite{Tack20_CSI}. 
We see from Table \ref{table_coreset} that just 2\% of the train\edit{ing }set is sufficient to achieve performance similar to the full train\edit{ing }set. In fact, for our clinical EEG task, we use just 2\% of the training samples for all experiments and reach the SOTA \edit{performance}. 

\begin{table}[h]
\captionsetup{skip=0pt}
\caption{\small Coreset \edit{Analyses}. AUROC for various coreset sizes (\% of training samples). Even 2\% of training data gives good results.}
\label{table_coreset}
\begin{center}
\begin{tabular}{@{}llrrrl@{}}
\toprule
                       & Coreset (\%) & \multicolumn{1}{l}{Audio} & \multicolumn{1}{l}{CIFAR} & \multicolumn{1}{l}{MNIST} & EEG                      \\ \midrule
\multirow{4}{*}{AUROC} & 1\%          & 73.60                      & 85.43                     & 96.90                      & -                        \\
                       & 2\%          & 73.85                     & 85.81                     & 97.02                     & \multicolumn{1}{r}{77.10} \\
                       & 10\%         & 73.90                      & 86.20                      & 97.10                      & -                        \\
                       & 100\%        & 73.99                     & 86.63                     & 97.21                     & -                        \\ \bottomrule
\end{tabular}
\end{center}
\vskip -0.2in
\end{table}

\subsection{Encoder capacity}
\label{results_encoder_supplementary}

\begin{figure}[h]
\captionsetup{belowskip=-15pt}
\begin{center}
\includegraphics[width=\textwidth/2]{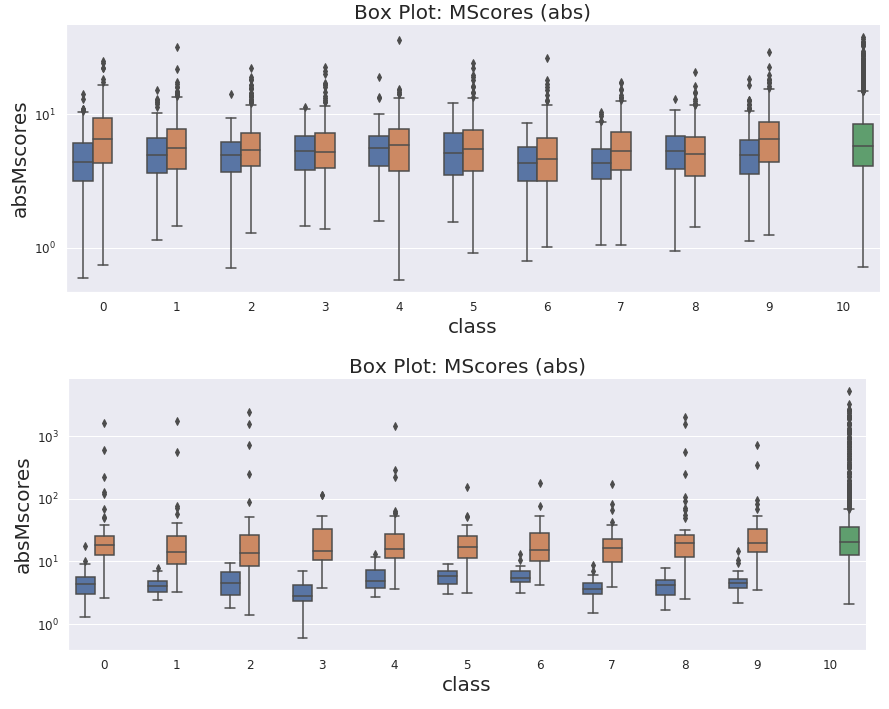}
\caption{\small Orange: GSC 0-9, Blue: FSDD 0-9, Green: \edit{All GSC-Words merged together}. 
(top) Low capacity Kymatio model trained on GSC 0-9. TRUST-LAPSE \edit{scores} cannot flag GSC-Words since scores overlap, indicating necessity of high capacity encoders. (bottom) Kymatio model trained on FSDD 0-9. TRUST-LAPSE scores \edit{will} flag GSC 0-9 examples, indicating \edit{the} model cannot generalize to GSC 0-9. \edit{This} is indeed true: accuracy on \edit{FSDD 0-9 is 0.97, accuracy on} GSC 0-9 is 0.42.}
\label{encoder_capacity_figure}
\vskip -0.1in
\end{center}
\vspace{-4mm}
\end{figure}

\textbf{Performance depends on encoder capacity.} We compare two different encoder architectures for our audio task, trained and evaluated identically: (i) a lower capacity encoder, Kymatio \cite{Andreux18_kymatio} (InD test set accuracy 0.75)\edit{;}  and (ii) the higher capacity M5 (Section \ref{experimental_setup}) encoder (InD test set accuracy 0.85). 
We observe that TRUST-LAPSE performs worse with Kymatio. It is not able to flag OOD examples (GSC-Words) based on TRUST-LAPSE scores (Fig. \ref{encoder_capacity_figure}, top). On the other hand, TRUST-LAPSE performs well with the M5 encoder (Table \ref{table_main-results}) and is able to correctly separate out semantic InD samples from semantic OOD samples. Thus, the success of  TRUST-LAPSE is highly dependent on the capacity of the trained encoder. 

\textbf{TRUST-LAPSE detects lack of generalization.} 
To simulate the setting where a model doesn't generalize well beyond the test set, we train\edit{ed} Kymatio for the same task (i.e., same InD semantic classes 0-9) but with the much smaller FSDD dataset. We see that it reaches 0.97 classification accuracy on the FSDD test set. On evaluation with GSC 0-9, it only \edit{achieves} 0.42 accuracy, showing it does not generalize well. 
\textbf{If we did \edit{not} have access to labelled GSC 0-9 examples (as is common in real-world settings), can TRUST-LAPSE identify this lack of generalization?} 
Indeed, we find that TRUST-LAPSE scores for this encoder  flag GSC 0-9 examples (Fig. \ref{encoder_capacity_figure}, bottom), indicating  that the model overfit to FSDD and would not have generalized to GSC 0-9. 

\section{Qualitative Explainability Evaluations}
\label{explainability_evaluation}

TRUST-LAPSE scores indicate when model predictions should and should not be trusted. 
They provide insight about model behavior when provided with a new input, guiding future actions (accept, flag, abstain, etc.). As seen in Section \ref{results_encoder_supplementary}, TRUST-LAPSE informs model generalizability and model error. 
Here, we describe (a) how TRUST-LAPSE scores can be explained, and (b) how trust scores can themselves be explanations used by practitioners during deployment. 

\begin{figure}[h]
\begin{center}
\includegraphics[width=\textwidth/2]{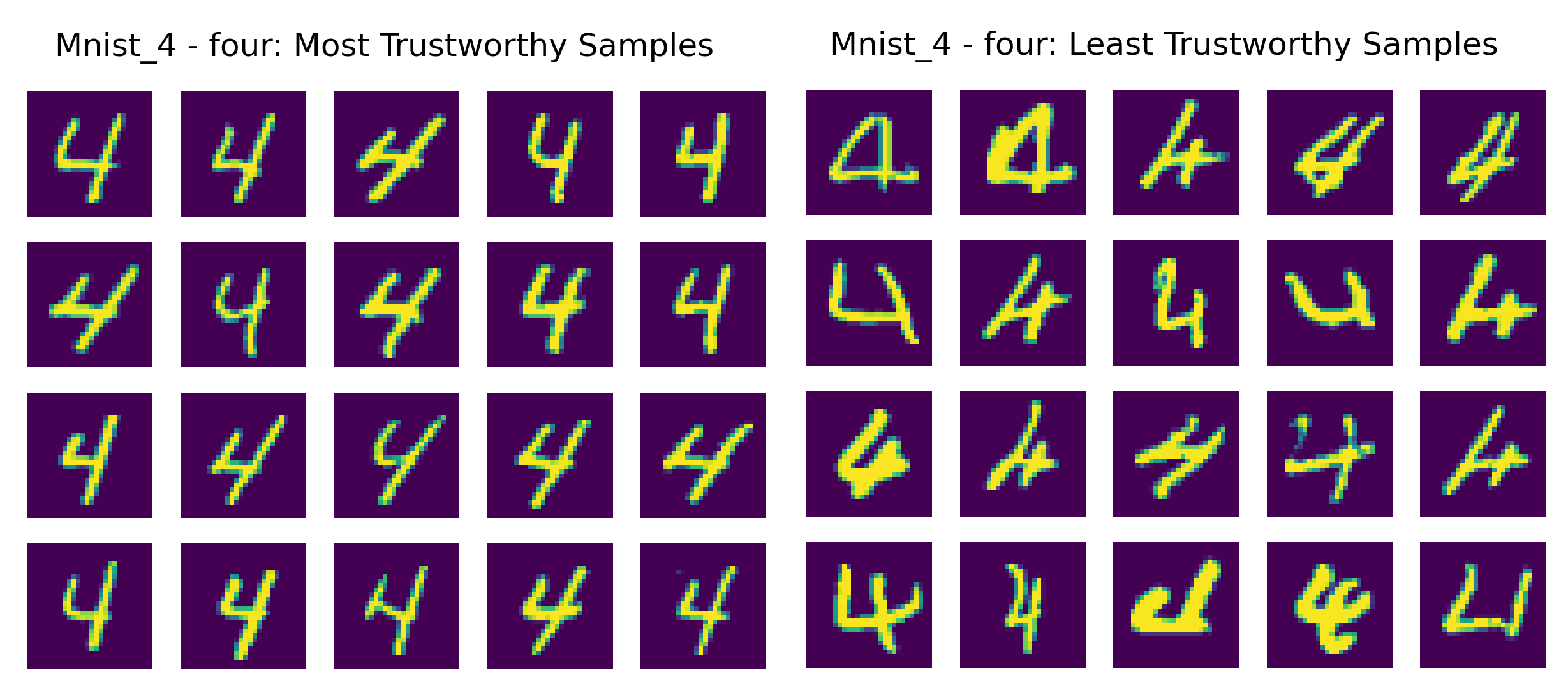}
\caption{\small MNIST Digit 4. (top left) New input. (left) Closest coreset samples to input, with high trust scores (right) Farthest coreset samples with low trust scores. More visualizations in Appendix, Figs. \ref{trust_score_examples_mnist_0}-\ref{trust_score_examples__cifar10_d}.}
\label{trust_score_examples_mnist_4}
\vskip -0.1in
\end{center}
\end{figure}

\textbf{Visualizing nearest coreset samples in the latent space.} To compute latent-space mistrust scores, TRUST-LAPSE uses coreset samples that are \textit{``closest''} according to distance metric $\textit{d}$ and similarity metric $\textit{sim}$. 
Thus, for any new sample, a practitioner can easily compare it with the closest and farthest coreset instances to explain TRUST-LAPSE's recommendation. For instance, Fig. \ref{trust_score_examples_mnist_4} shows samples from MNIST class 4. For a new input (top left), visualizing the closest and farthest coreset samples (in this case, with the highest and lowest trust scores) can explain why its prediction is trusted (with trust score 0.97), because the input sample resembles highly trusted samples in the coreset and is very different from low trust samples. 
In general, identifying ways in which a new sample is different from the coreset, both in the input space and in the latent-space, via such visualizations, can help explain its trust score. 

\begin{figure}[h]
\begin{center}
\includegraphics[width=\textwidth/2]{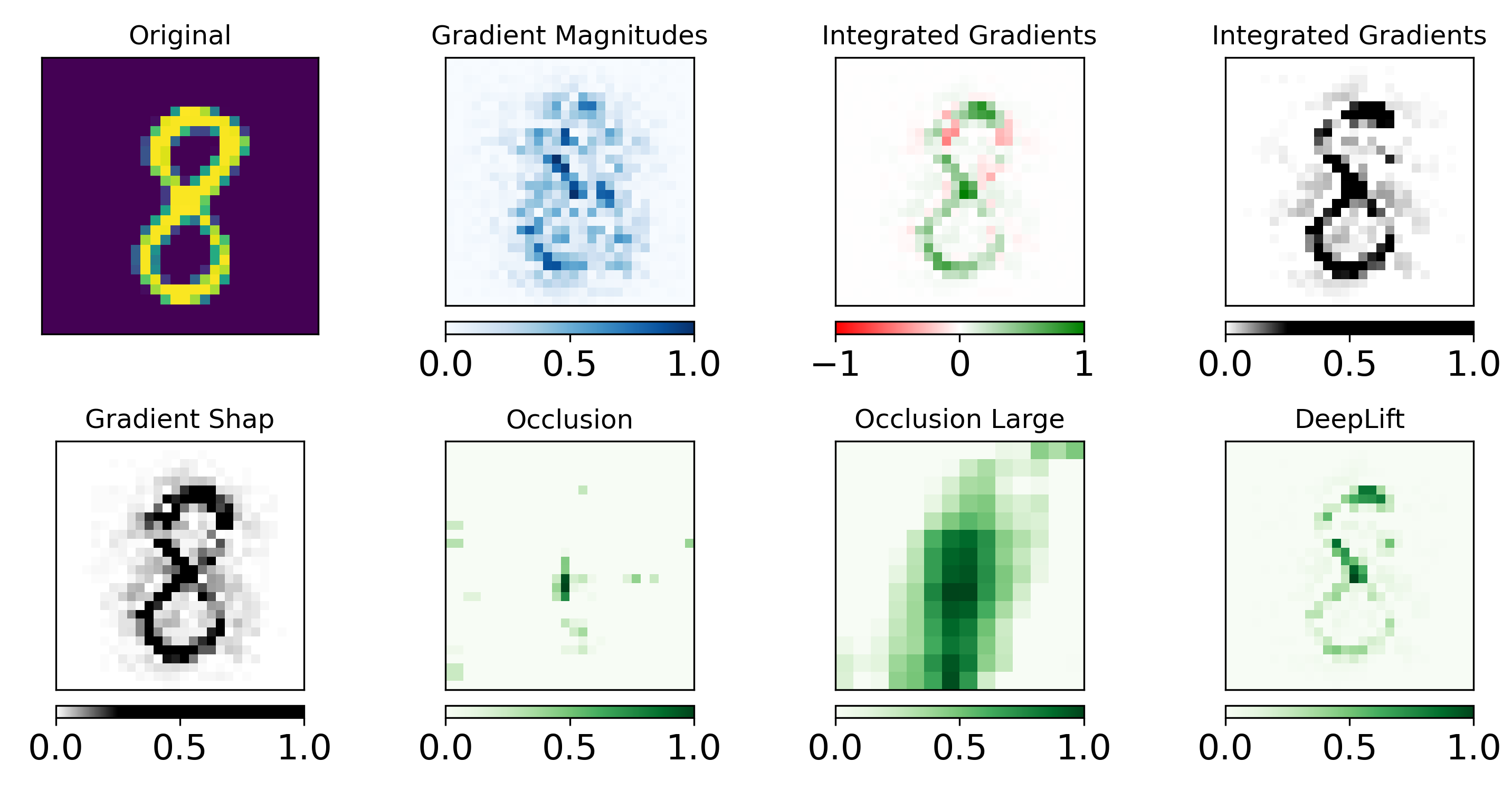}
\caption{\small MNIST Digit 8, Model prediction 8. TRUST-LAPSE score = 0.95 (high). High trust scores correlate to correct model predictions and good attributions from Integrated Gradients, DeepLIFT and GradientSHAP. Saliency and Occlusions are less useful.}
\label{attributions_mnist8_high}
\vskip -0.1in
\end{center}
\vspace{-6mm}
\end{figure}

\textbf{Using Attribution Methods to explain TRUST-LAPSE scores.} We use popular attribution methods such as Saliency \cite{Simoyan2013_saliency}, Integrated Gradients \cite{Sundararajan2017_integratedGrad}, DeepLift \cite{Shrikumar2017_DeepLift}, GradientSHAP \cite{lundberg2017_gradshap}, Occlusion \cite{zeiler2014_occlusion}, and variants of Layer-wise Relevance Propgation (LRP) \cite{kindermans2016_lrpProof} to analyse which parts of the input are most important for the model prediction (details in Appendix \ref{attribution_settings}). On manual analysis, we find that high trust scores correlate to correct model predictions and good attributions (Fig. \ref{attributions_mnist8_high}). Low trust scores, however, correlate to model errors\edit{,} and their attributions do not corresponding correctly to the class (Fig. \ref{attributions_mnist8_low}). More examples \edit{are} in Appendix, Figs. \ref{attributions_mnist0_high}-\ref{attributions_cifar_horse_high}, \edit{and} show that TRUST-LAPSE scores are inherently interpretable and can be explained together with any good attribution-based method.

\begin{figure}[h]
\begin{center}
\includegraphics[height=0.2\textheight]{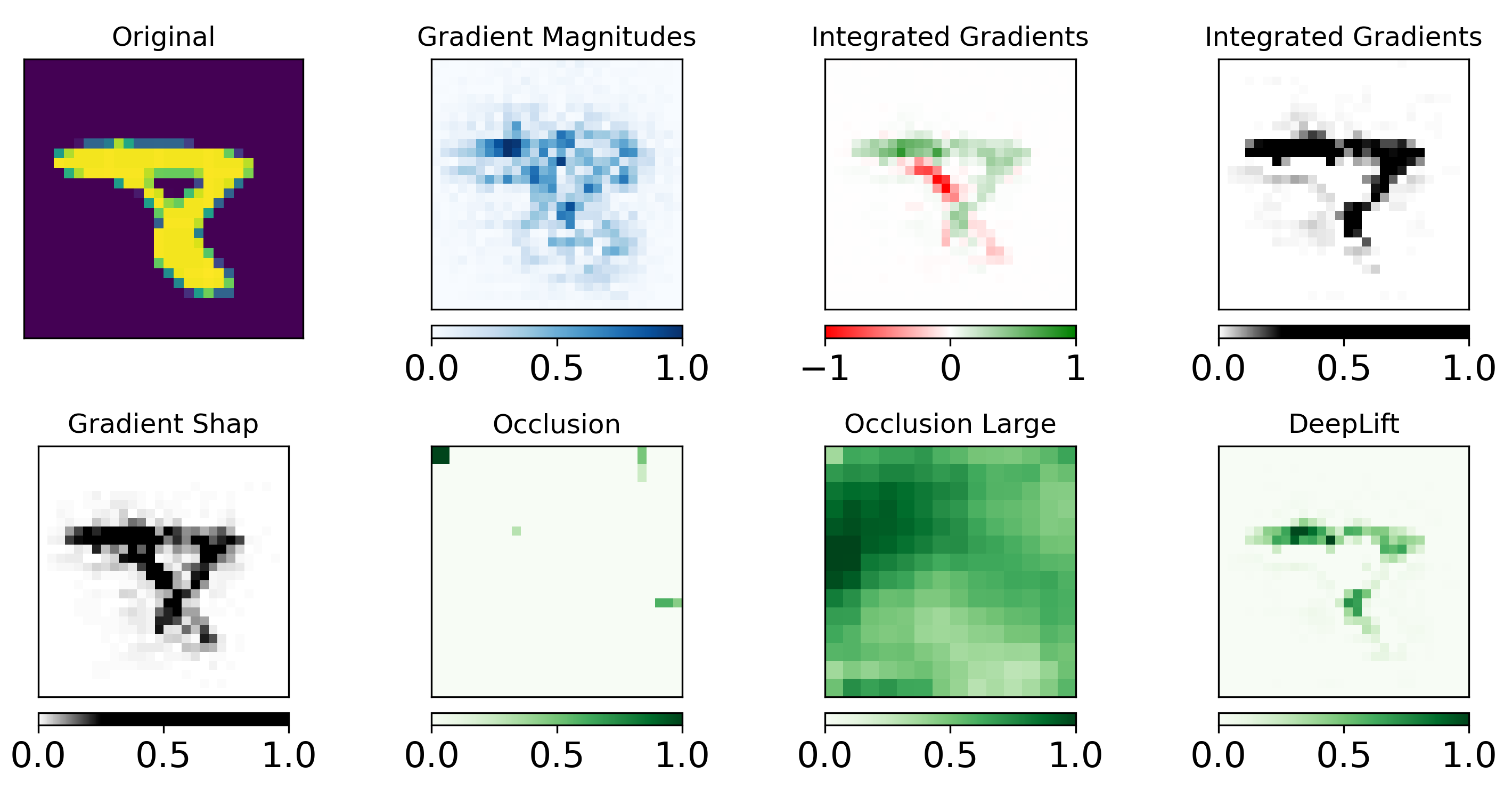}
\caption{\small MNIST Digit 8, Model prediction 7. TRUST-LAPSE score = 0.44 (low). Low trust scores correlate to model errors while corresponding attributions from Integrated Gradients, DeepLIFT and GradientSHAP show that highlighted portions of the image do not correspond correctly to the class label. Saliency (gradient magnitudes) and Occlusions are less useful.}
\label{attributions_mnist8_low}
\vskip -0.1in
\end{center}
\vspace{-8mm}
\end{figure}

\textbf{Visualizing latent-space embeddings.} Since the model's learnt latent-space is key for TRUST-LAPSE, we can visualize the latent-space embeddings of the coreset and incoming samples using UMAP \cite{umap} (Appendix, Figs.  \ref{gsc_umap}-\ref{eeg_umap}) to get insights behind the trust score. 
Visually examining UMAP projections of latent-space representations for the input in relation to those of the coreset can illustrate the cluster the input is most similar to, providing qualitative explanations. New inputs projected far from the class clusters are less trustworthy. 
Note that UMAP is not distance preserving and may be noisy.

\textbf{Plotting TRUST-LAPSE Scores for datastreams.} \edit{A} useful visualization comes from tracking various scores of TRUST-LAPSE (i.e., latent-space mistrusts score, sequential mistrust scores, etc.) across the datastream on a plot. A practitioner, having access to such plots on their dashboard,  updated over real-time, can easily deduce explanations regarding changing data distributions, model performance \edit{and} drift. 
For instance, in Figs. \ref{short_plots} \& \ref{audio_seq_100_main}, we see, at a glance, how input data distributions are changing just from the mistrust score plots and can verify by examining these samples in the input space.

\begin{figure}[h]
\begin{center}
\includegraphics[width=0.25\textwidth]{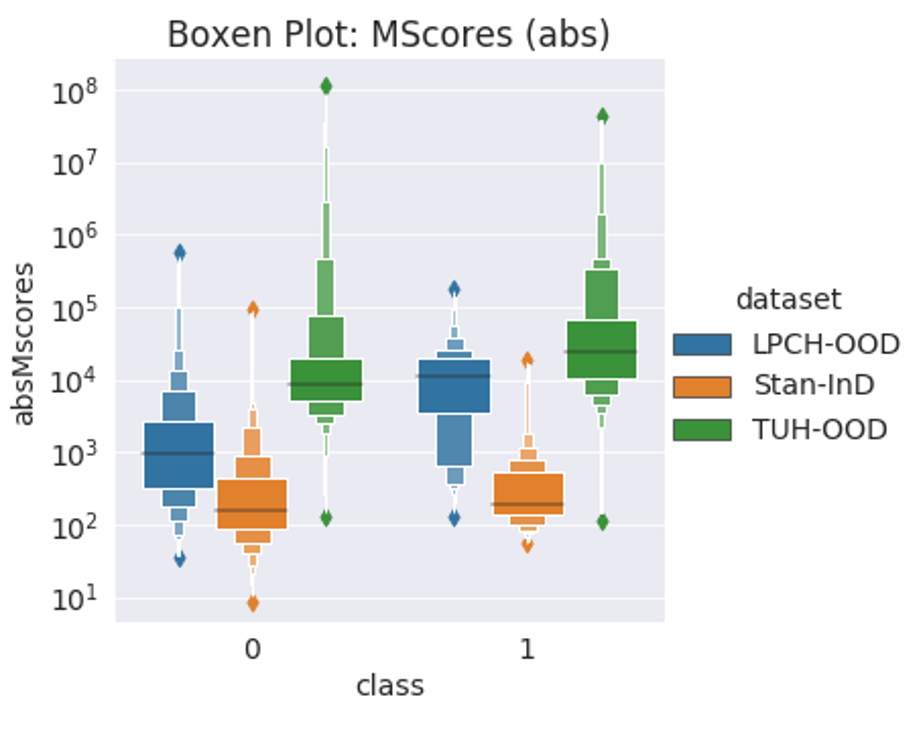}
\vskip -0.15in
\caption{\small Latent-Space Mistrust Score distributions on EEG data. }
\label{eeg_boxen}
\end{center}
\vspace{-8mm}
\end{figure}

\textbf{Analysing TRUST-LAPSE Score Distributions.} 
Plotting distributions of TRUST-LAPSE scores across different datasets (or if in real-time: sets or windows of data sampled from the datastream) can help explain characteristics of the data (Fig. \ref{eeg_boxen}). 
The plot suggests that 
LPCH data distributions (blue) are closer to SHC (orange) than TUH (green). This is indeed true: SHC and LPCH share similar hardware infrastructure and are located in the same area. 
They differ only in the ages of the patient population they serve. TUH, however, has lot of dissimilarities in hardware, patient populations, age group, etc.

\begin{figure}[h]
\begin{center}
\includegraphics[width=\textwidth/2]{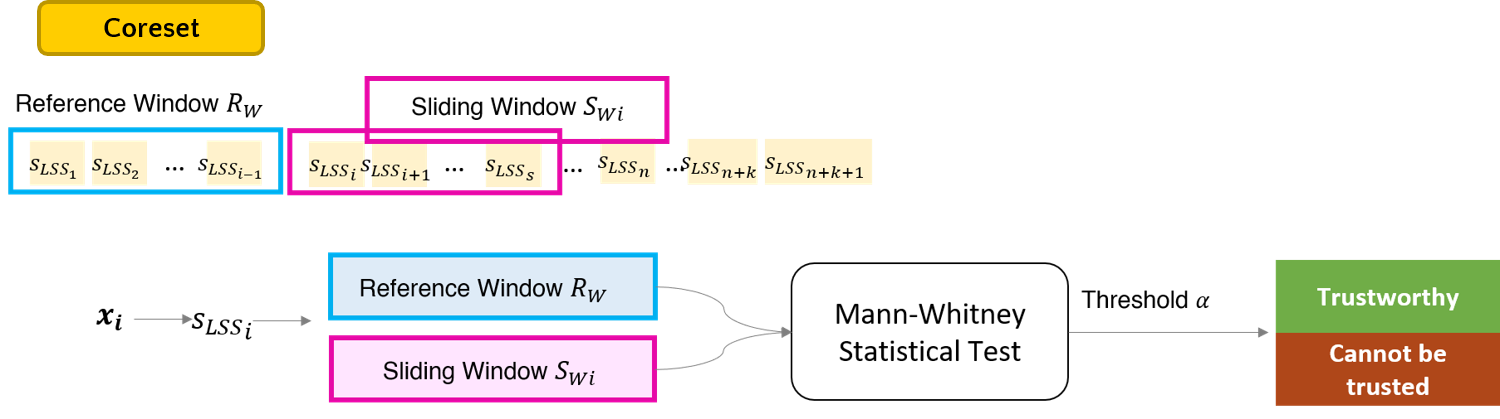}
\caption{\small Sequential Mistrust Scoring. A reference window $R_w$ is drawn from the latent-space mistrust scores of the coreset and the sliding window $S_{w_i}$ from those of the input data stream. }
\label{seq}
\vskip -0.1in
\end{center}
\vspace{-6mm}
\end{figure}

\textbf{Sequential Aspect: Samples from the sliding window.} The sequential aspect of TRUST-LAPSE also gives the practitioner certain other tools to analyse and debug their ML system. Often in real-world applications, as time passes, samples start to drift from the original expected distribution (especially important in clinical settings) \cite{Gama14_surveyConceptDrift}. To compute sequential mistrust scores, TRUST-LAPSE utilizes a sliding window of a fixed window size over the last few samples and compares it with reference window sampled from the coreset (Fig. \ref{seq}). 
Thus, when TRUST-LAPSE raises an alert, a practitioner can look at the sliding window as a whole and compare it with the reference window (in terms of inputs, latent-space embeddings \edit{and} scores) to identify drifts that are  noticeable only by looking at a window of samples. 

\section{Discussion}
\label{discussion}

\textbf{Coreset \& Training Data.} TRUST-LAPSE requires access to trained models along with a small fraction of training data (Section \ref{results_coreset_size}). However, in most real settings, training data may not be available. In such cases, data samples that the model is known to perform well on can act as the coreset. Our coreset \edit{analyses} also paves the way for choosing different data sampling strategies that can ensure more ``representational" samples (based on domain knowledge). 

\textbf{Mitigation opportunities.} In case the practitioner finds that a model should have performed well on certain samples (trustworthy predictions) but they are being flagged incorrectly by TRUST-LAPSE, mitigation is \edit{often} easy. 
The practitioner can simply examine the coreset, add few more representative samples to it or remove bad samples from it. Similarly, the practitioner can also update samples in the reference window to improve sequential mistrust scoring. In fact, they can choose to observe the scores with \textbf{\textit{several}} reference windows to see patterns and differences between the flagged samples and the reference samples. This puts control of TRUST-LAPSE fully into the hands of the practitioner. Thus, not only can we explain trust scores and use TRUST-LAPSE for explaining, but can very easily improve it based on domain knowledge.

\textbf{Generalization.} Ideally, high-performing models that are well-trained, rigorously tested and validated ought to generalize well to new environments, populations, and scenarios. However, in reality, they may not do so without needing fine-tuning or retraining. Our preliminary experiments (Section \ref{results_encoder_supplementary}) suggest that accurate mistrust scoring from TRUST-LAPSE can pinpoint when the model needs to be fine-tuned or retrained. 

\textbf{Flexibility.} TRUST-LAPSE can accept any type of distance and similarity metrics as well as sequential scoring measures. We investigated several classical metrics (Section \ref{results_individual_scores}) and verified that our sequential mistrust and latent-space mistrust are complementary. 
This suggests that there may be more metrics, even learnt ones, which can be easily incorporated into TRUST-LAPSE. Moreover, our sequential formulation provides an opportunity to leverage contextual information present in a set or sequence of inputs which cannot be harnessed by standard approaches that look at samples in isolation. It is independent of latent-score mistrust and can be applied on any score sequence.

\section{Conclusions \& Future Work}
\label{conclusions}
We develop a mistrust scoring framework, TRUST-LAPSE, that can quantify trustworthiness of model predictions during inference and perform continuous model monitoring. We use the model's own latent-space representations along with a robust sequential, sliding window test to inform trust in model predictions. Our framework is explainable (each metric and score is human interpretable), post-hoc (can work with any trained model), actionable (automated and provides concrete decisions) and high-performing. 
We achieve SOTA on distributionally shifted input detection with our approach on three diverse domains -- vision, audio and clinical EEGs. Furthermore, we achieve high drift detection rates, crucial for continuous model monitoring. Through our extensive experiments and analyses, we expose critical flaws in current works, provide an explainability workflow and understand where performance benefits come from. We hope TRUST-LAPSE will be adopted for routinely characterizing model trust and model performance in the wild. We will investigate (i) developing alternative, learnt metrics for our latent-space mistrust\edit{;} (ii) using our sequential windowing approach on other scores apart from latent-space mistrust\edit{;}
and (iii) using our mistrust scores for active learning in future work.

\section*{Acknowledgements}
We would like to thank Ashwin Paranjape, Siyi Tang, Khaled Saab, Florian Dubost, \edit{members of the Chaudhari Lab,} and all anonymous reviewers for feedback on the manuscript that
has significantly improved this work.

\section*{Code \& Data Availability}
All our code for TRUST-LAPSE, baselines and all the experiments presented in the paper \edit{are} publicly available on GitHub \edit{at \url{https://github.com/nanbhas/TrustLapse}}. All audio and vision datasets as well as the TUH datasets are publicly
available (Section \ref{experimental_setup}). SHC and LPCH data contain patient sensitive information and cannot be shared publicly. Details on data preprocessing steps, experimental settings provided in Appendix \ref{experimental_setup-details}.

\bibliographystyle{ieeetr} 
\bibliography{refs}

\clearpage
\setcounter{page}{1}
\section*{Appendix}
\label{appendix}

\subsection{Detailed Experimental Setup}
\label{experimental_setup-details}

\textbf{Image Encoders \& Pre-processing} We \edit{performed} our vision experiments using CNN-based classifiers, i.e. the LeNet architecture \cite{Lecun1998_mnistLeNet} when working with $x$-MNIST data and the ResNet18 architecture \cite{He16_resnet} when working with CIFAR10/100/SVHN data. With the $x$-MNIST data, we  normalize\edit{d} the inputs using the mean and standard deviation calculated from the training set for both training and inference, as is standard. With the CIFAR10/100/SVHN data, we appl\edit{ied} the normalization transform along with augmentation strategies (random crop and horizontal flips) during training and only normalization during inference. $x$-MNIST data have dimensionality $28 \times 28 = 784$ while CIFAR-like data have dimensionality $32 \times 32 \times 3 = 3072$.

\textbf{Audio Encoders \& Pre-processing} For our audio experiments, we use\edit{d} the M5 classifier \cite{Dai16_m5} as our encoder. We 
also train\edit{ed} a simple two-layer network -- a static, normalized, log-scattering layer that extracts scattering coefficients from audio signals \cite{Andreux18_kymatio} followed by a log-softmax layer that generates output probabilities to compare encoder capabilities (Section \ref{results_encoder_supplementary}). Both models are trained only on classes 0-9 from the GSC dataset with the rest of the classes as OOD inputs. In all cases, we learn\edit{ed} embeddings directly from raw, one second audio clips (resampled to 8kHz) without making use of any spectrogram features like MFCC \cite{mfcc}. Raw audio data have dimensionality of $1 \times 8000 = 8000$, much larger than the vision datasets.

\textbf{EEG Encoders \& Pre-processing} For our seizure detection task, we use\edit{d} the Dense-Inception based models \cite{Saab20_DenseInceptionEEG} as encoders. We use\edit{d} the same data pre-processing strategy as given by \cite{Saab20_DenseInceptionEEG}. The models are trained on 19 channel, 12 second or 60 second raw EEG clips from InD datasets, 
resampled to 200Hz. EEG clips have dimensionality of $60 \times 200 \times 19 = 228,000$, much larger than audio or vision datasets.

\textbf{Additional Details} We train\edit{ed} all encoders on InD training data using standard weight initializations, SGD/Adam optimizer and ReLU non-linearities. We train\edit{ed} and tune\edit{d} hyperparameters for our encoders using only InD data samples. We \edit{did not} expose any outlier data to our encoders. We extract\edit{ed} activations from a fully connected (FC) hidden layer before the logits layer to form latent-space representations. 
We normalize\edit{d} our distance scores and similarity scores before forming the latent-space mistrust scores in cases where one of them is tightly bound while the other is not (for instance, cosine similarity is bound to $[-1,1]$). For our sequential mistrust scoring, we set window sizes $w_A,\, w_B = 25$ for vision tasks and $w_A, \, w_B = 50$ for audio and EEG tasks (Remark \ref{window_size}). We chose the reference window by sampling randomly from the training set (or coreset) and using its latent-space mistrust scores. For generating the data streams for drift detection, we randomly shuffle\edit{d} the \edit{test} set and \edit{chose} data points sequentially. We first obtain\edit{ed} the latent-space mistrust scores for the \edit{test} set forming a 1D evaluation sequence, on which we \edit{performed} the sliding window analysis sequentially. 

\subsection{Supplementary Tables}

Table \ref{table_main_significance_image} shows performance on CIFAR100 and $x$-MNIST datasets which have partial semantic overlap between train distribution and test distributions, scores are over 5 random runs. In our MNIST experiments, we see  TRUST-LAPSE performs comparably with Test-Time Dropout. It has the best AUPR and comparable  AUROC and FPR80 values. When CIFAR100 classes are added (naively considered disjoint from CIFAR10 as in standard benchmarks) to the evaluation, TRUST-LAPSE shows the best AUPR score and the second best AUROC and FPR80.

\begin{table*}[h]
\captionsetup{skip=0pt}
\caption{\small OOD performance: CIFAR10 image classification on $x$-MNIST and CIFAR100 which have partial semantic overalp. Mean and standard deviations over five random runs. Best scores indicated in bold.}
\label{table_main_significance_image}
\begin{center}
\begin{small}
\begin{tabular}{@{}lcccccc@{}}
\toprule
                     & \multicolumn{3}{c}{$x$-MNIST}                                                         & \multicolumn{3}{c}{CIFAR100}                                 \\ \cmidrule(l){2-7} 
\multicolumn{1}{c}{} & AUROC $\uparrow$    & AUPR $\uparrow$   & FPR80 $\downarrow$                     & AUROC $\uparrow$    & AUPR $\uparrow$   & FPR80 $\downarrow$ \\ \midrule
MSP                  & 0.899 $\pm$ 0.006 & 0.979   $\pm$ 0.002 & 0.148   $\pm$ 0.012 & 0.852   $\pm$ 0.014 & 0.986 $\pm$ 0.001 & 0.231 $\pm$ 0.018  \\
Predictive Entropy   & 0.902 $\pm$ 0.006 & 0.981 $\pm$ 0.002   & 0.147 $\pm$ 0.011 & 0.854   $\pm$ 0.014 & 0.986 $\pm$ 0.001 & 0.230 $\pm$ 0.017 \\
KL\_U                & 0.899 $\pm$ 0.007 & 0.980 $\pm$ 0.002   & 0.164 $\pm$ 0.012  & 0.860   $\pm$ 0.015 & 0.987 $\pm$ 0.001 & 0.223 $\pm$ 0.023  \\
ODIN                 & 0.898 $\pm$ 0.006 & 0.979 $\pm$ 0.002   & 0.148 $\pm$ 0.015 & 0.845   $\pm$ 0.018 & 0.986 $\pm$ 0.001 & 0.251 $\pm$ 0.031  \\
Vanilla Mahalanobis  & 0.918 $\pm$ 0.006 & 0.984 $\pm$ 0.001   & 0.118 $\pm$ 0.015 & 0.679   $\pm$ 0.012 & 0.967 $\pm$ 0.001 & 0.558 $\pm$ 0.025 \\
Test-Time Dropout    & \textbf{0.976 $\pm$ 0.002} & 0.986 $\pm$ 0.001   & \textbf{0.016 $\pm$ 0.001} & \textbf{0.925   $\pm$ 0.004} & 0.986 $\pm$ 0.000 & \textbf{0.049 $\pm$ 0.003}  \\
 TRUST-LAPSE (ours)    & \textbf{0.972 $\pm$ 0.002} & \textbf{0.995 $\pm$ 0.000}   & 0.037 $\pm$ 0.038 & 0.866   $\pm$ 0.008 & \textbf{0.988 $\pm$ 0.001} & 0.239 $\pm$ 0.028  \\ \bottomrule
\end{tabular}
\end{small}
\end{center}
\vskip -0.1in
\end{table*}

\subsection{Attribution Algorithm Settings}
\label{attribution_settings}
Pytorch Captum \cite{kokhlikyan2020captum} was used for all attribution algorithm implementations. Default settings were used for Saliency, Integrated Gradients, DeepLift and GradientSHAP. For GradientSHAP, a random image distribution was provided as baseline while a zero image was provided as baselines for Integrated Gradients and DeepLIFT. For Occlusion and Occlusion large, window sizes of $(5,5)$ and $(10,10)$ with stride $(2,2)$ were used respectively. We do not show Layer-wise Relevance Propagation (LRP) attributions explicitly since it has been shown that for networks that have only ReLU or maxpooling operations, the z-rule of LRP is equivalent to the gradient multiplied by the input when biases are included as inputs and the epsilon value used for numerical stability is set to 0. This equivalence is shown in references \cite{Shrikumar2017_DeepLift} and \cite{kindermans2016_lrpProof}. All networks used in this paper have only ReLU or maxpooling operations and thus, the equivalence holds true.

\subsection{Additional Figures and Visualizations}

\subsubsection{\textbf{Drift Detection}}

\begin{figure*}[h]
\includegraphics[width=\textwidth]{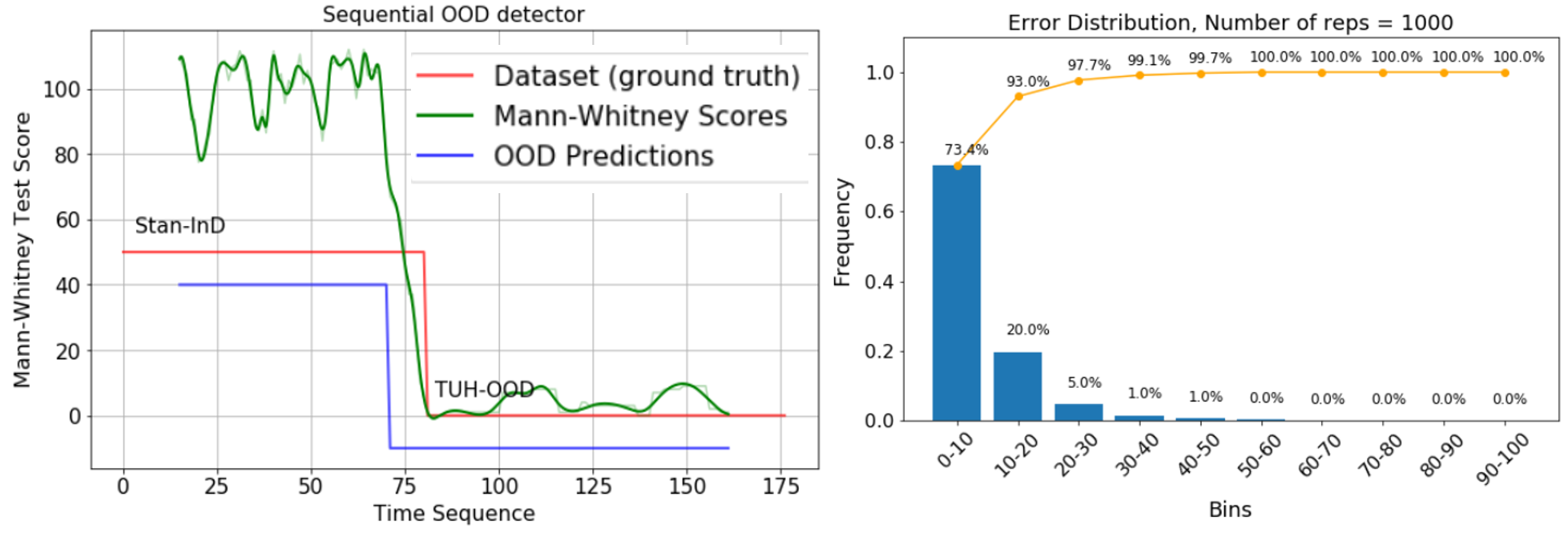}
\caption{\small    (left) Snapshot of the  TRUST-LAPSE mistrust scores. Within a window of 15 samples, it is able to detect the change point for the seizure detection task. (right) \% Error distribution over 1000 trials.}
\label{sequentialOOD_results}
\end{figure*}

\begin{figure*}[ht]
\includegraphics[width=\textwidth]{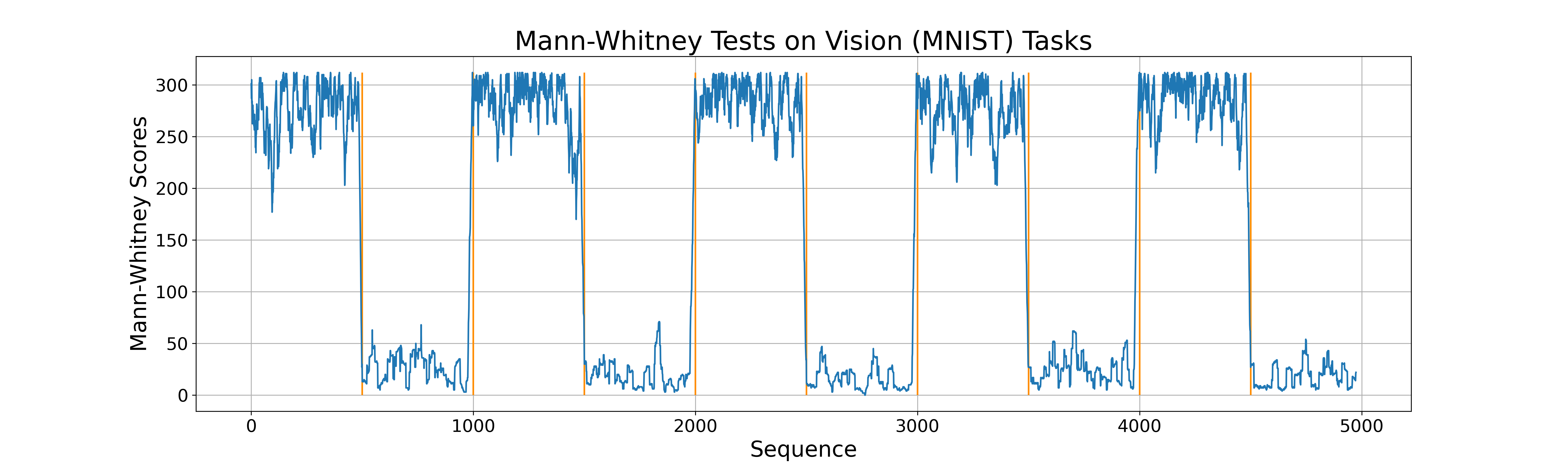}
\caption{\small    Drift Detection (MNIST): Data stream with distribution shifts occurring every 500 steps. The orange lines indicate true change points. Blue plot represents the sequential mistrust scores..}
\label{mnist_seq}
\end{figure*}

\begin{figure}[h]
\begin{center}
\includegraphics[width=\textwidth/2]{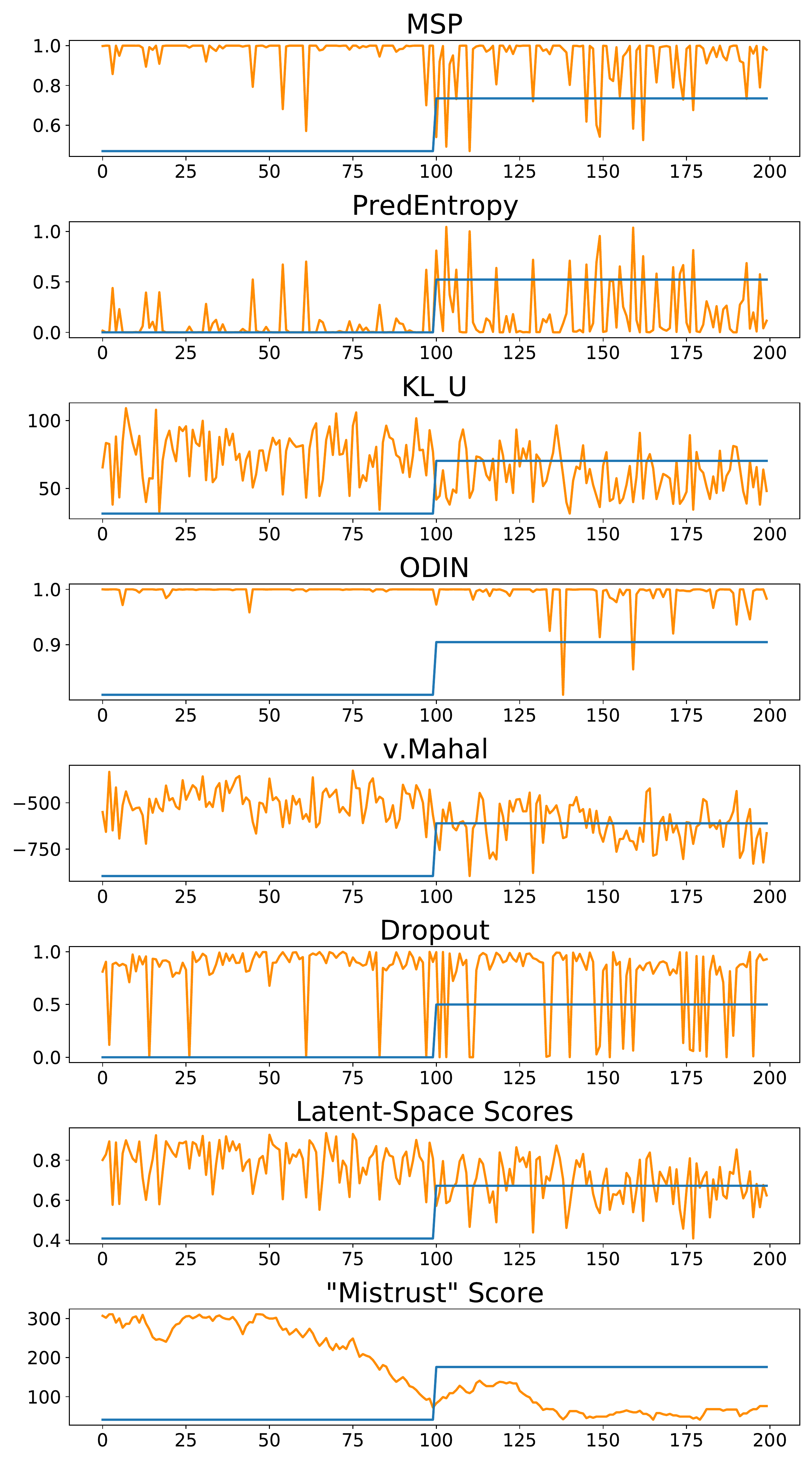}
\caption{\small Trust scores from Baselines and TRUST-LAPSE (CIFAR).  Sequential Mistrust Score shows the most separability between trustworthy (blue low segment) and to be flagged (blue high segment) points.}
\label{sequential_ablations_cifar}
\end{center}
\end{figure}

\begin{figure}[h]
\begin{center}
\includegraphics[width=\textwidth/2]{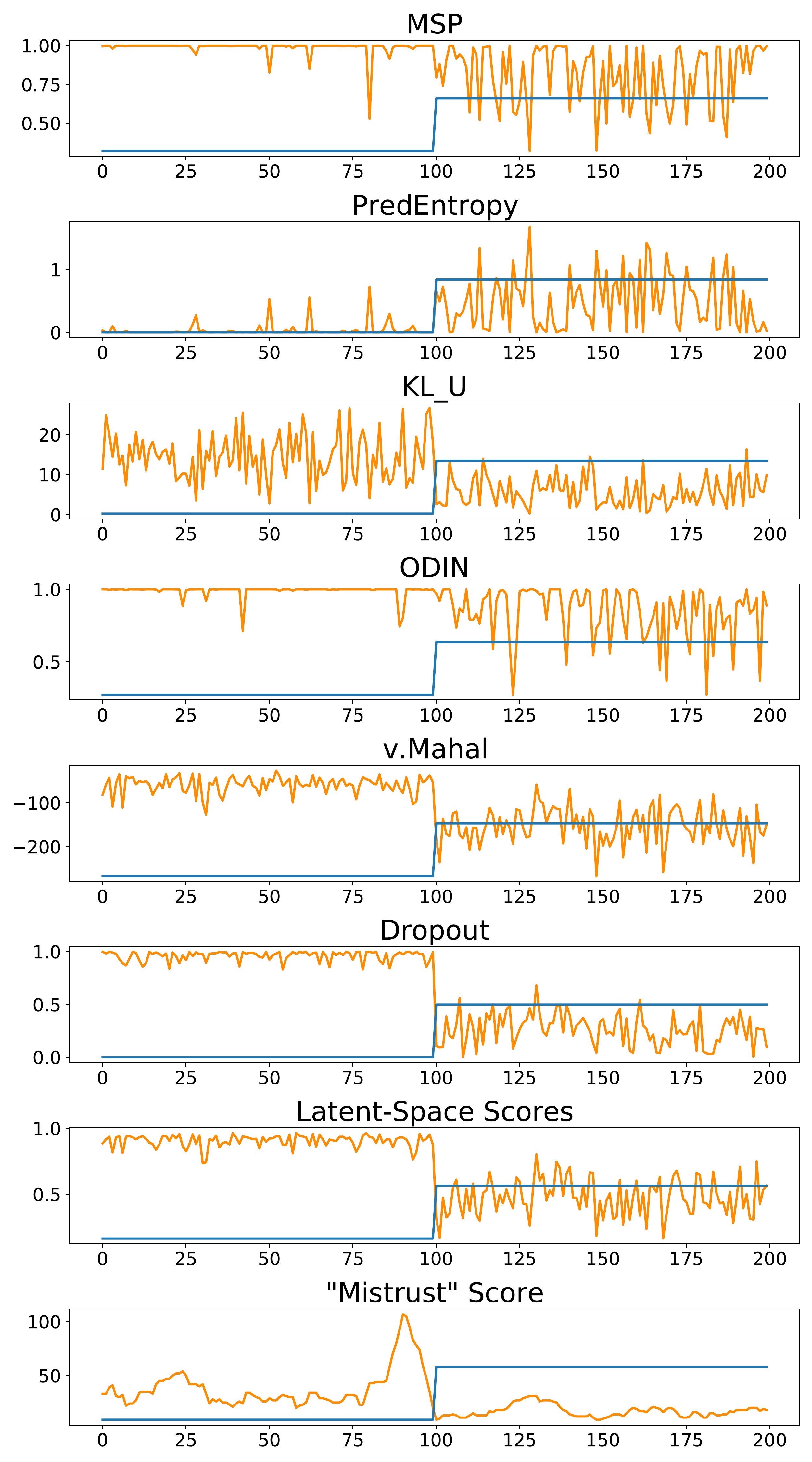}
\caption{\small Trust scores from Baselines and TRUST-LAPSE (MNIST).  Sequential Mistrust Score shows the most separability between trustworthy (blue low segment) and to be flagged (blue high segment) points.}
\label{sequential_ablations_mnist}
\end{center}
\end{figure}

\begin{figure*}[ht]
\includegraphics[width=\textwidth]{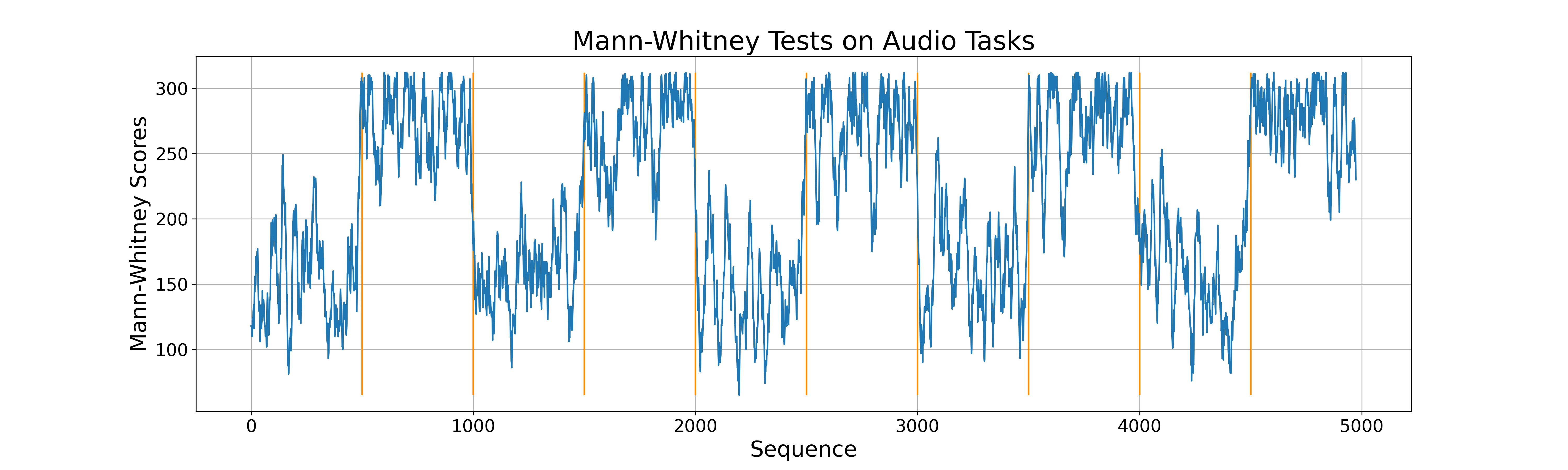}
\caption{\small    Drift Detection: Data stream with distribution shifts occurring every 500 steps. The orange lines indicate true change points. Blue plot represents the sequential mistrust scores}
\label{audio_seq_500}
\end{figure*}

\begin{figure*}[ht]
\includegraphics[width=\textwidth]{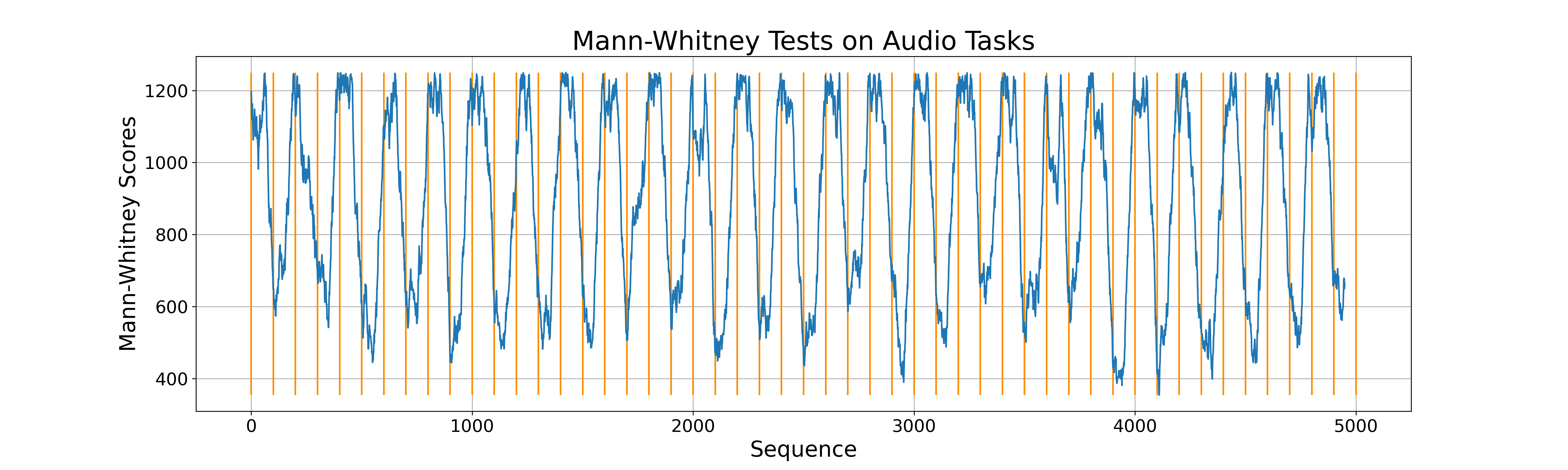}
\caption{\small    Drift Detection: Data stream with distribution shifts occurring every 100 steps. The orange lines indicate true change points. Blue plot represents the sequential mistrust scores}
\label{audio_seq_100}
\end{figure*}

\begin{figure*}[ht]
\includegraphics[width=\textwidth]{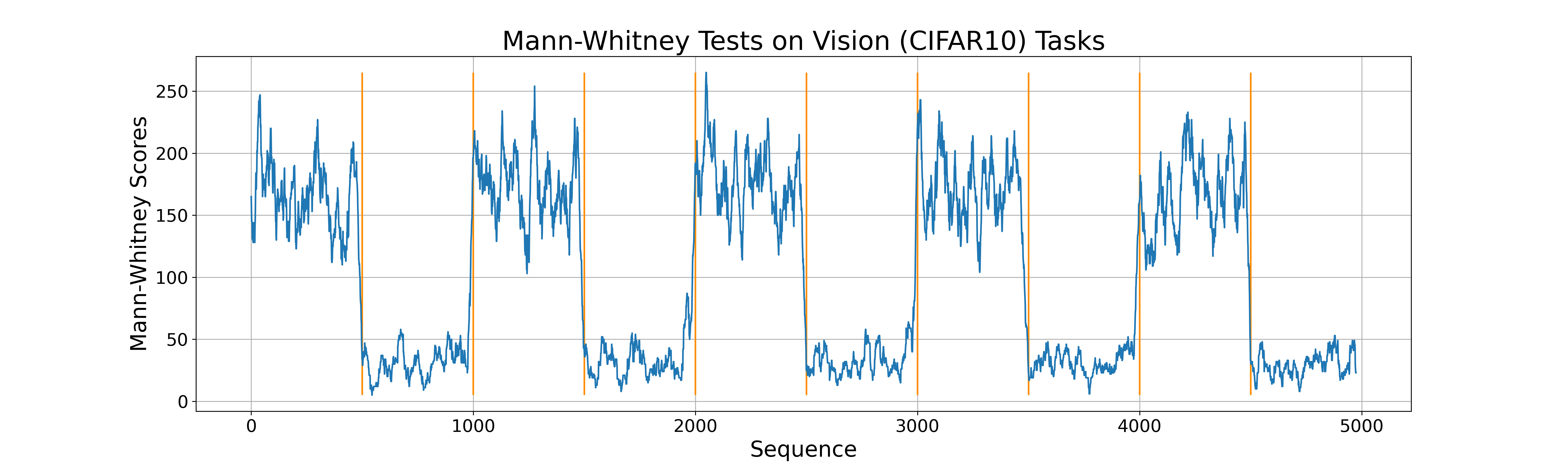}
\caption{\small    Drift Detection (CIFAR): Data stream with distribution shifts occurring every 500 steps. The orange lines indicate true change points. Blue plot represents the sequential mistrust scores.}
\label{cifar_seq}
\end{figure*}

Some examples of data streams generated from sampling EEG, audio and vision datasets are shown in Figs. \ref{sequentialOOD_results}, \ref{mnist_seq} \ref{sequential_ablations_cifar}, \ref{sequential_ablations_mnist}, \ref{audio_seq_500}, \ref{audio_seq_100}, and \ref{cifar_seq}). The The X axis indicates the time sequence. The Y axis gives the scores (Mann Whitney Scores are the same as  mistrust scores). Using a two cluster KMeans algorithm, change points are detected and predictions on individual samples are trusted or flagged.\\

\subsubsection{\textbf{Visualizing nearest coreset samples in the latent space}} 
\begin{figure}[h]
\begin{center}
\includegraphics[width=\textwidth/2]{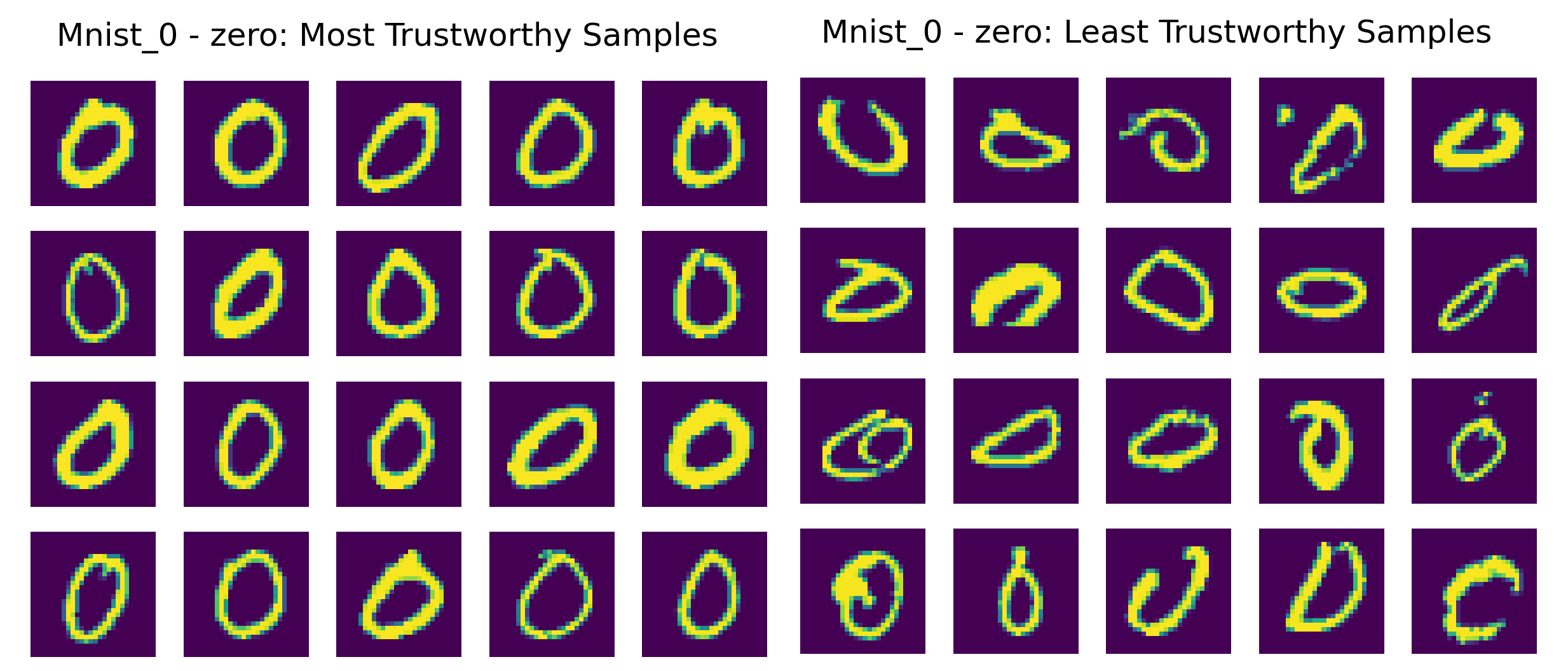}
\caption{\small MNIST Digit 0. (top left) New input. (left) Closest coreset samples to input, with high trust scores (right) Farthest coreset samples with low trust scores.}
\label{trust_score_examples_mnist_0}
\vskip -0.1in
\end{center}
\end{figure}

\begin{figure}[h]
\begin{center}
\includegraphics[width=\textwidth/2]{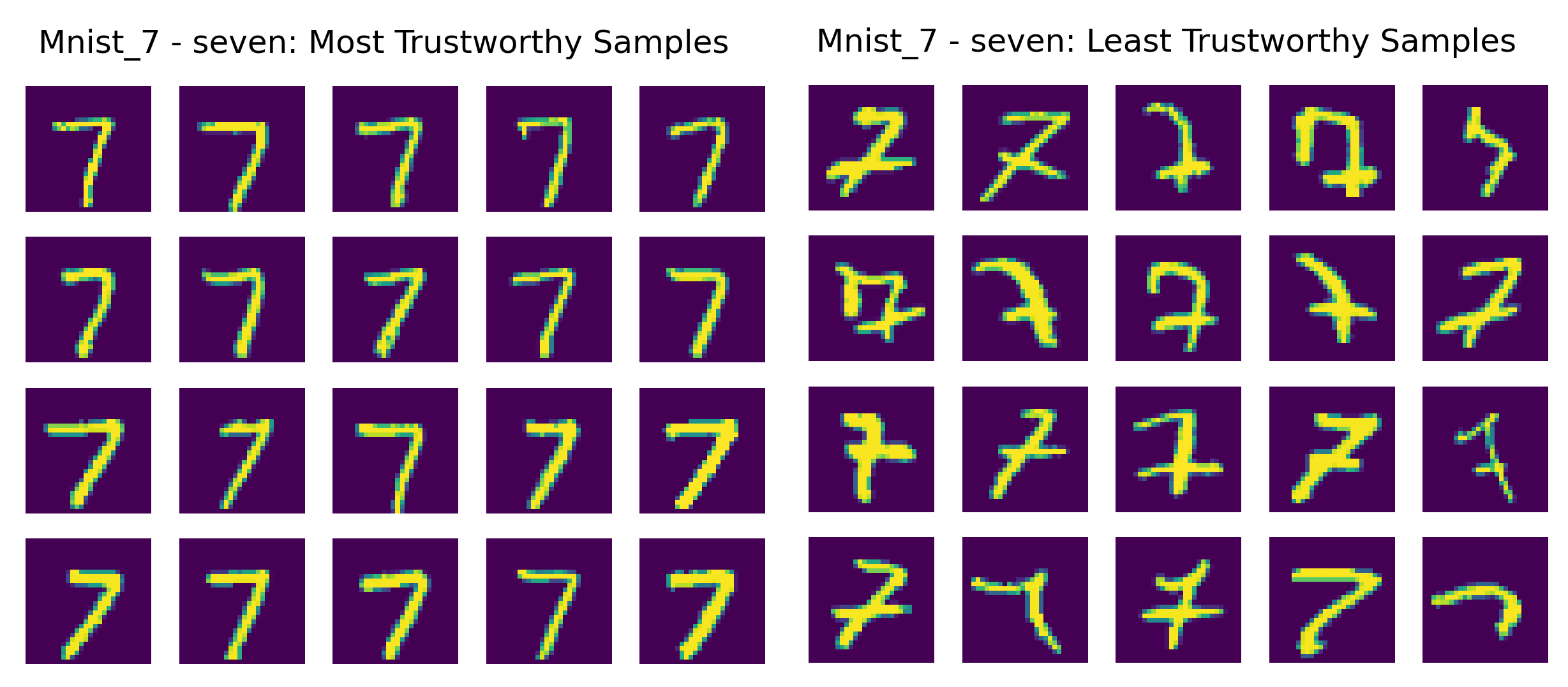}
\caption{\small MNIST Digit 7. (top left) New input. (left) Closest coreset samples to input, with high trust scores (right) Farthest coreset samples with low trust scores.}
\label{trust_score_examples_mnist_7}
\vskip -0.1in
\end{center}
\end{figure}

\begin{figure}[h]
\begin{center}
\includegraphics[width=\textwidth/2]{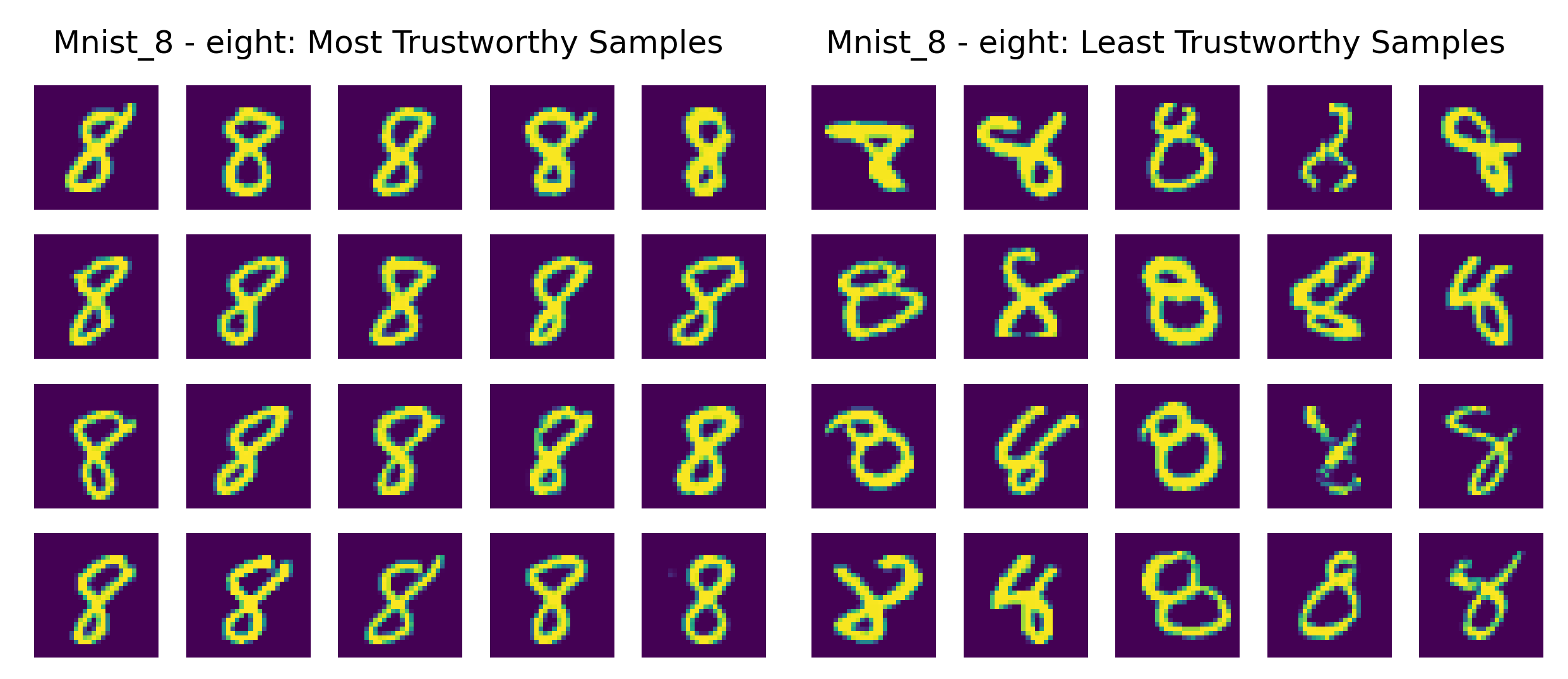}
\caption{\small MNIST Digit 8. (top left) New input. (left) Closest coreset samples to input, with high trust scores (right) Farthest coreset samples with low trust scores.}
\label{trust_score_examples_mnist_8}
\vskip -0.1in
\end{center}
\end{figure}

\begin{figure}[h]
\begin{center}
\includegraphics[width=\textwidth/2]{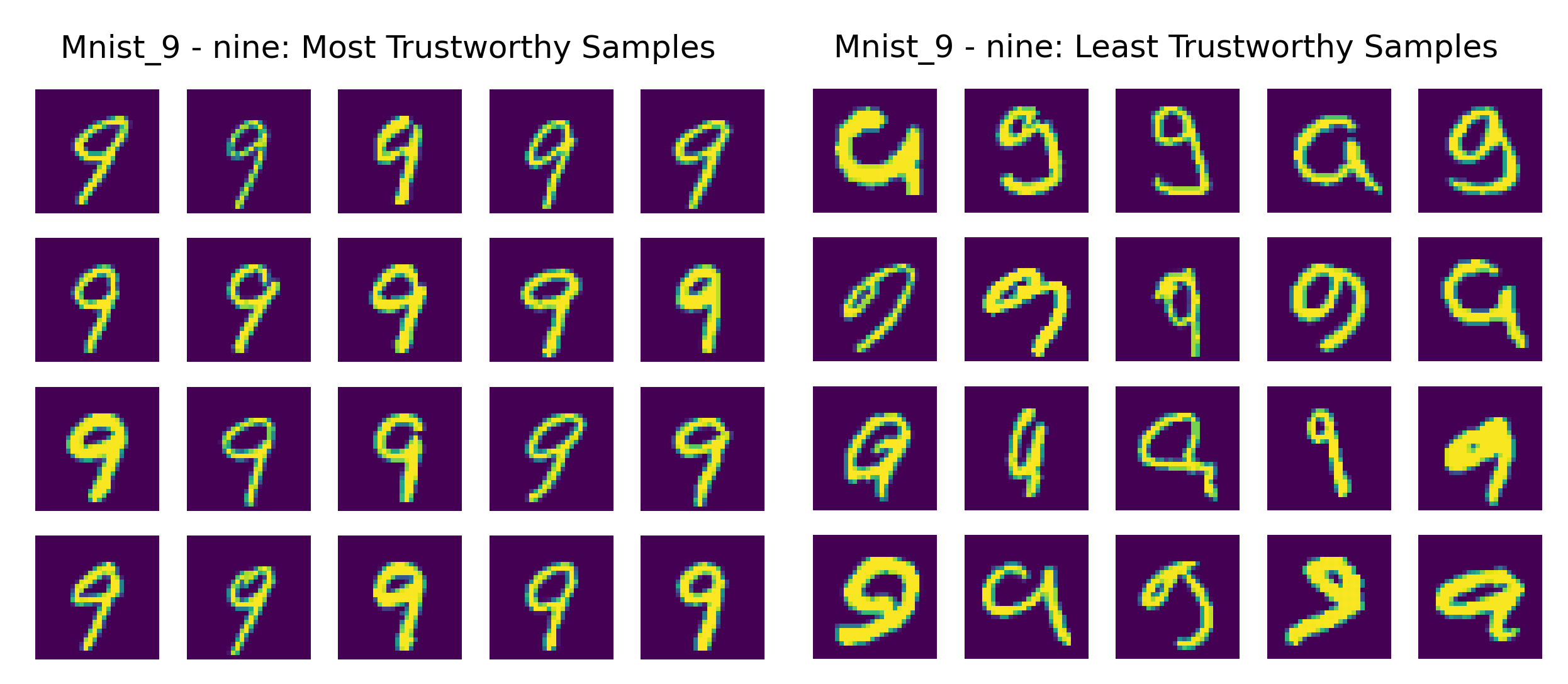}
\caption{\small MNIST Digit 9. (top left) New input. (left) Closest coreset samples to input, with high trust scores (right) Farthest coreset samples with low trust scores.}
\label{trust_score_examples_mnist_9}
\vskip -0.1in
\end{center}
\end{figure}

\begin{figure}[h]
\begin{center}
\includegraphics[width=\textwidth/2]{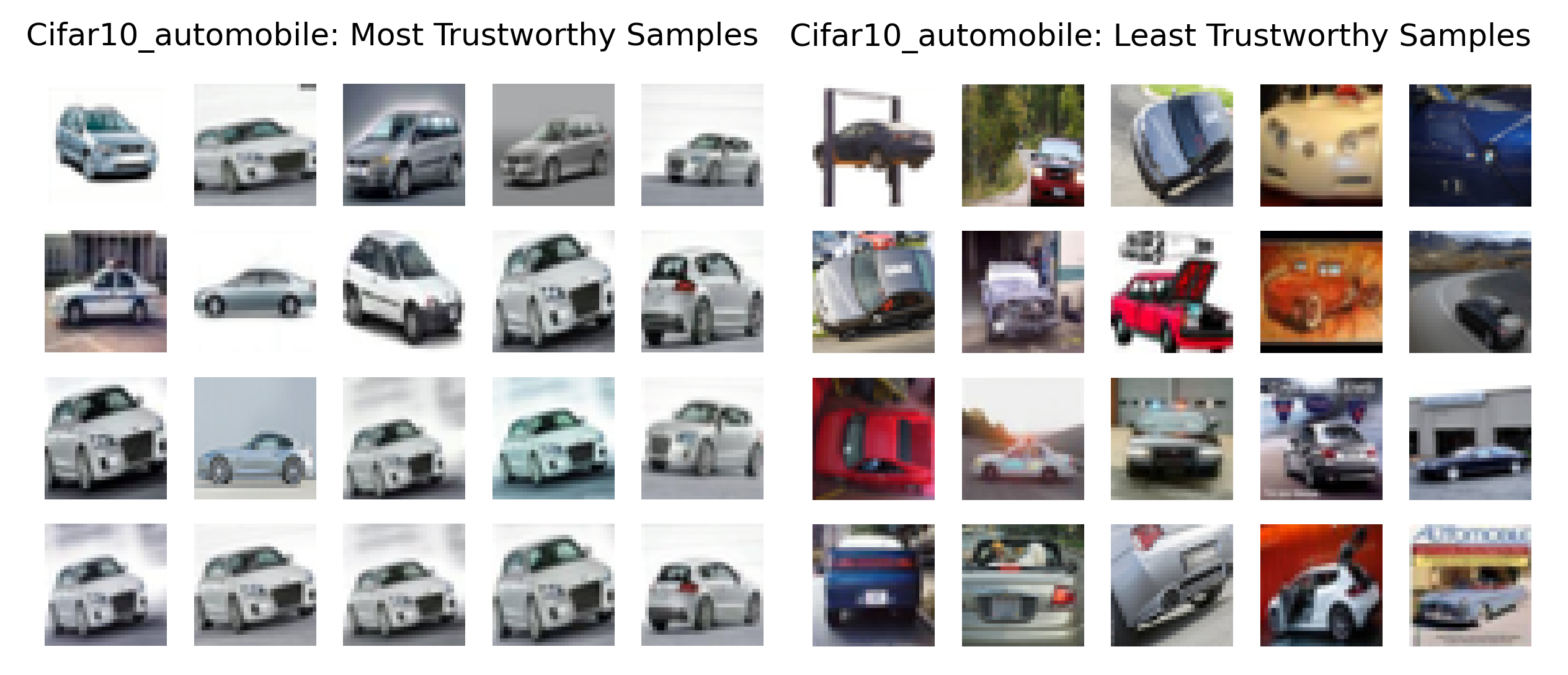}
\caption{\small CIFAR10 Automobile. (top left) New input. (left) Closest coreset samples to input, with high trust scores (right) Farthest coreset samples with low trust scores.}
\label{trust_score_examples_cifar10_a}
\vskip -0.1in
\end{center}
\end{figure}

\begin{figure}[h]
\begin{center}
\includegraphics[width=\textwidth/2]{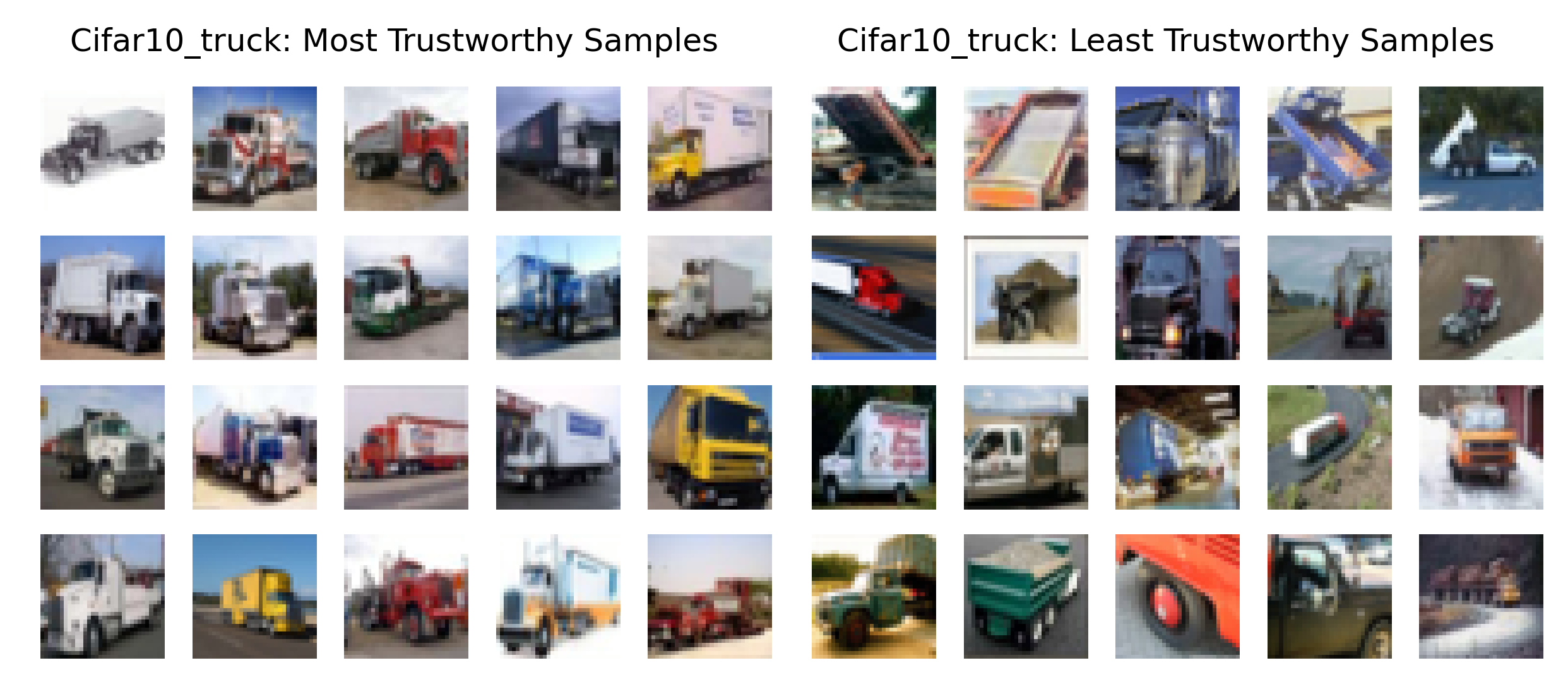}
\caption{\small CIFAR10 Truck. (top left) New input. (left) Closest coreset samples to input, with high trust scores (right) Farthest coreset samples with low trust scores.}
\label{trust_score_examples_cifar10_b}
\vskip -0.1in
\end{center}
\end{figure}

\begin{figure}[h]
\begin{center}
\includegraphics[width=\textwidth/2]{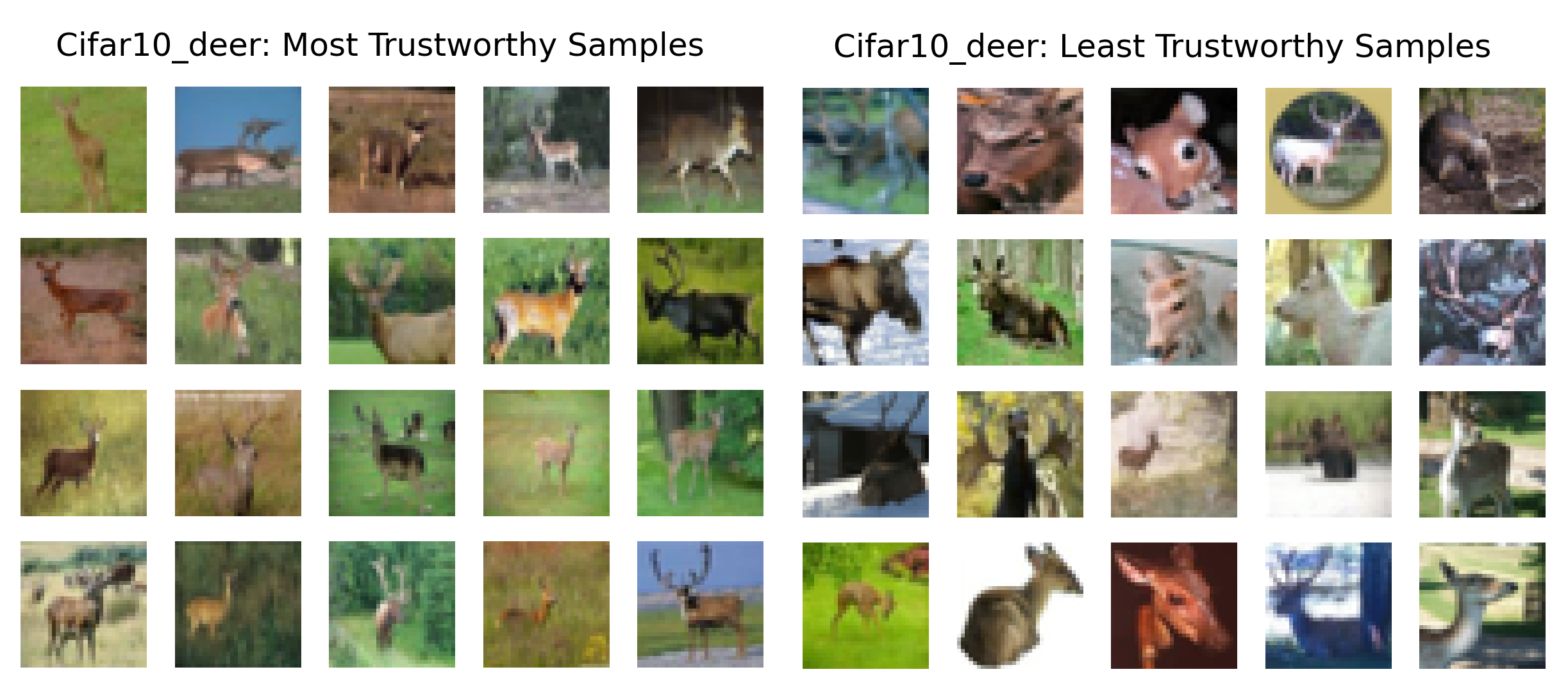}
\caption{\small CIFAR10 Deer. (top left) New input. (left) Closest coreset samples to input, with high trust scores (right) Farthest coreset samples with low trust scores.}
\label{trust_score_examples_cifar10_c}
\vskip -0.1in
\end{center}
\end{figure}

\begin{figure}[h]
\begin{center}
\includegraphics[width=\textwidth/2]{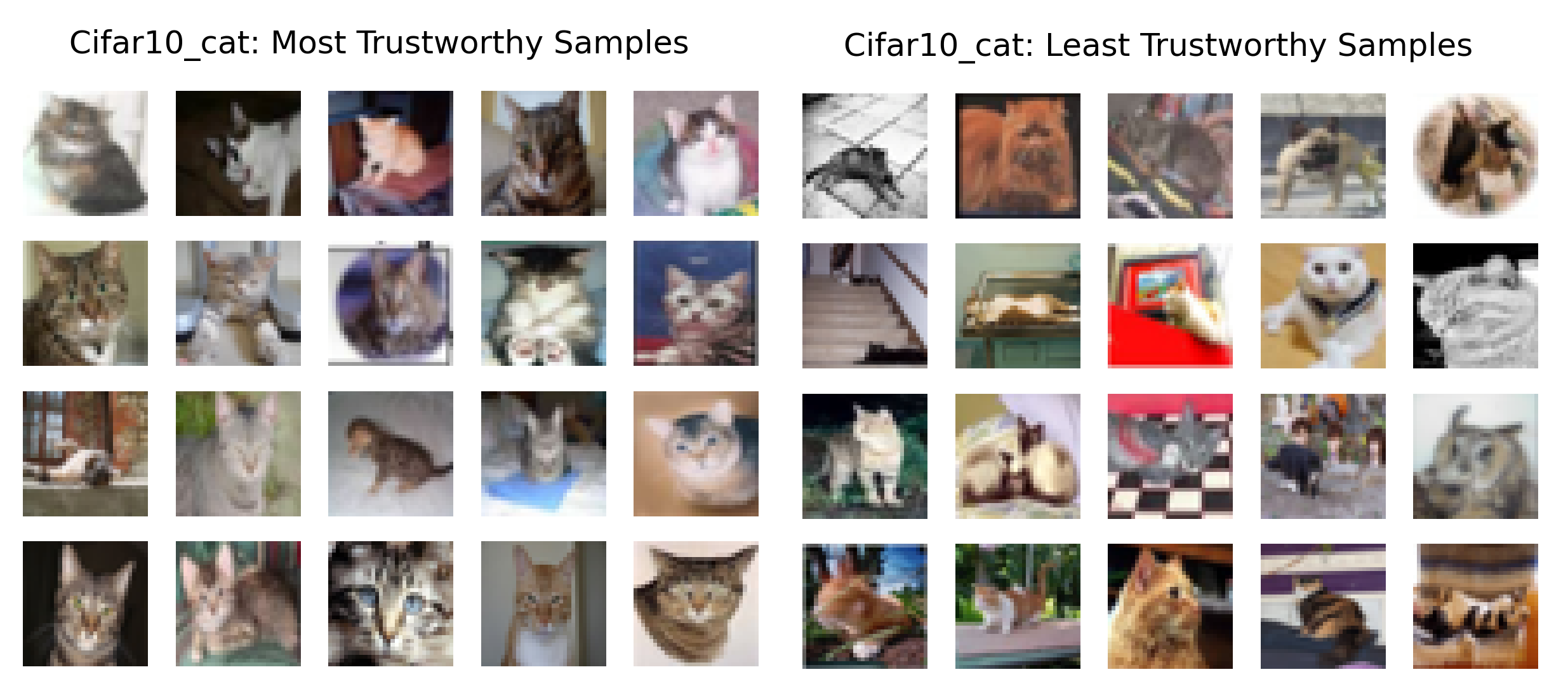}
\caption{\small CIFAR10 Cat. (top left) New input. (left) Closest coreset samples to input, with high trust scores (right) Farthest coreset samples with low trust scores.}
\label{trust_score_examples__cifar10_d}
\vskip -0.1in
\end{center}
\end{figure}

Additional class-wise examples of most trustworthy and least trustworthy samples are given in Figs. \ref{trust_score_examples_mnist_0} - \ref{trust_score_examples__cifar10_d}.\\

\subsubsection{\textbf{Using Attribution Methods to explain TRUST-LAPSE scores}} 
\begin{figure}[h]
\begin{center}
\includegraphics[width=\textwidth/2]{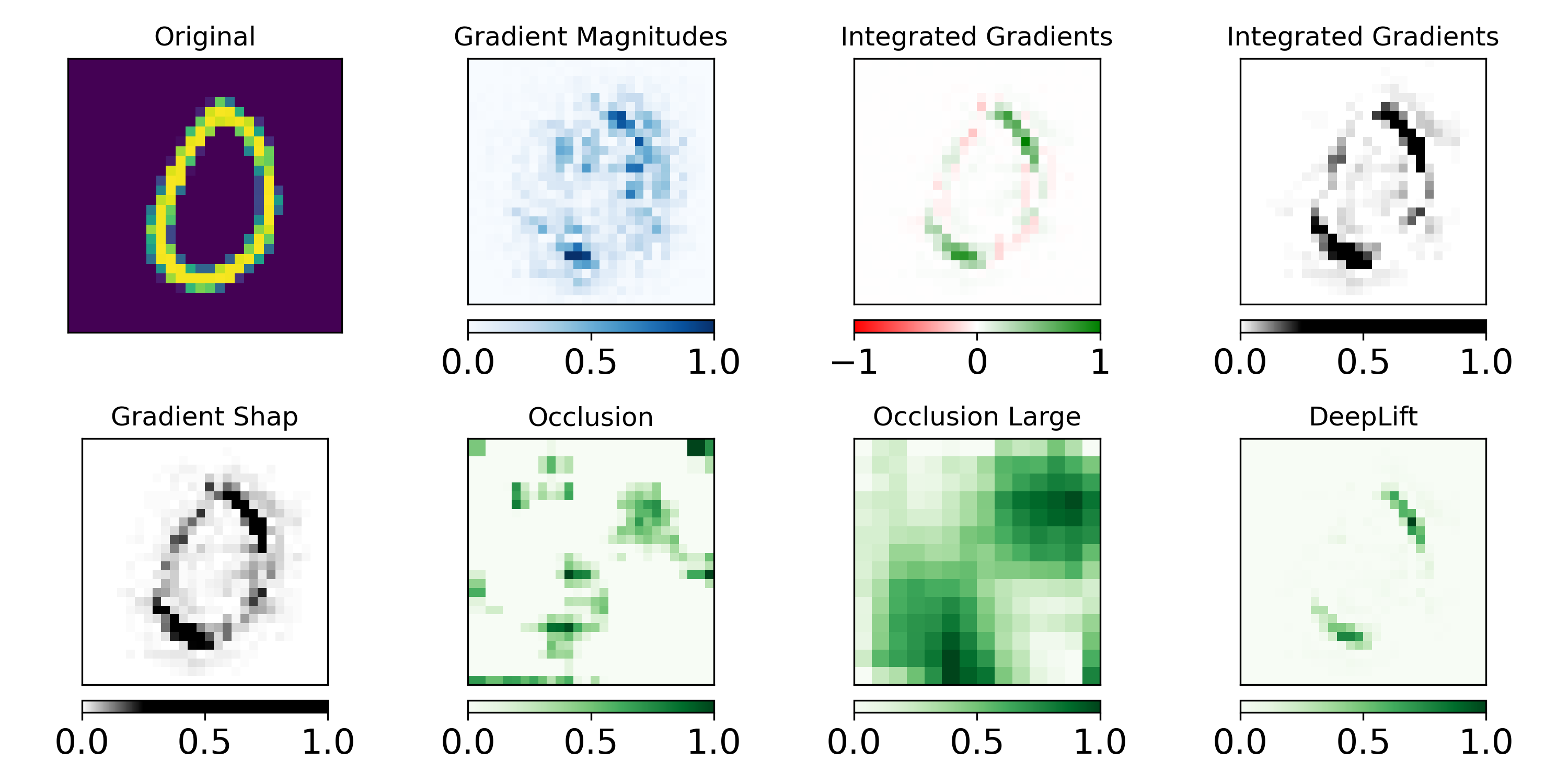}
\caption{\small MNIST Digit 0, Model prediction 0. TRUST-LAPSE score = 0.97 (high).}
\label{attributions_mnist0_high}
\vskip -0.1in
\end{center}
\end{figure}

\begin{figure}[h]
\begin{center}
\includegraphics[width=\textwidth/2]{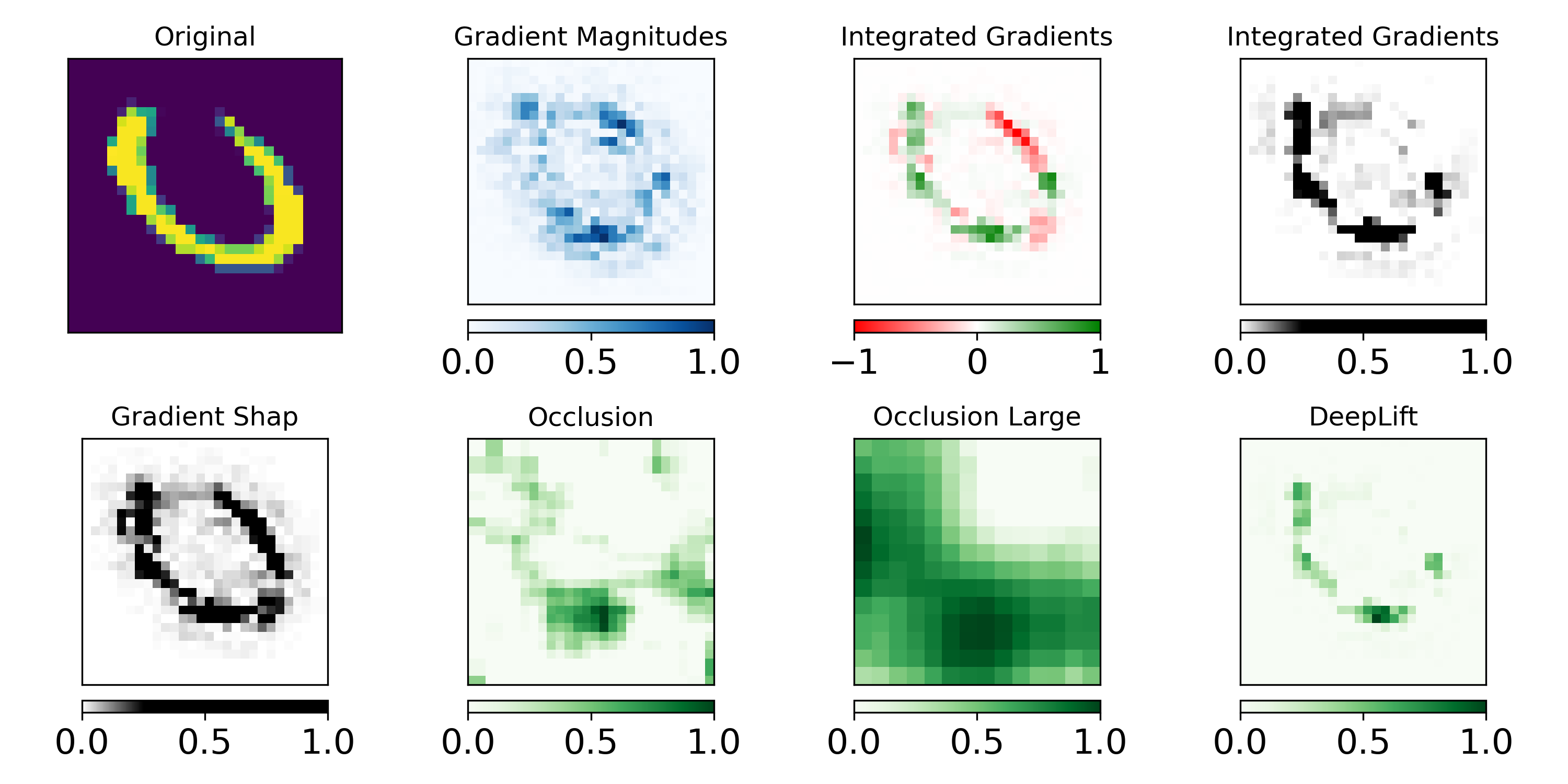}
\caption{\small MNIST Digit 0, Model prediction 6. TRUST-LAPSE score = 0.45 (low).}
\label{attributions_mnist0_low}
\vskip -0.1in
\end{center}
\end{figure}

\begin{figure}[h]
\begin{center}
\includegraphics[width=\textwidth/2]{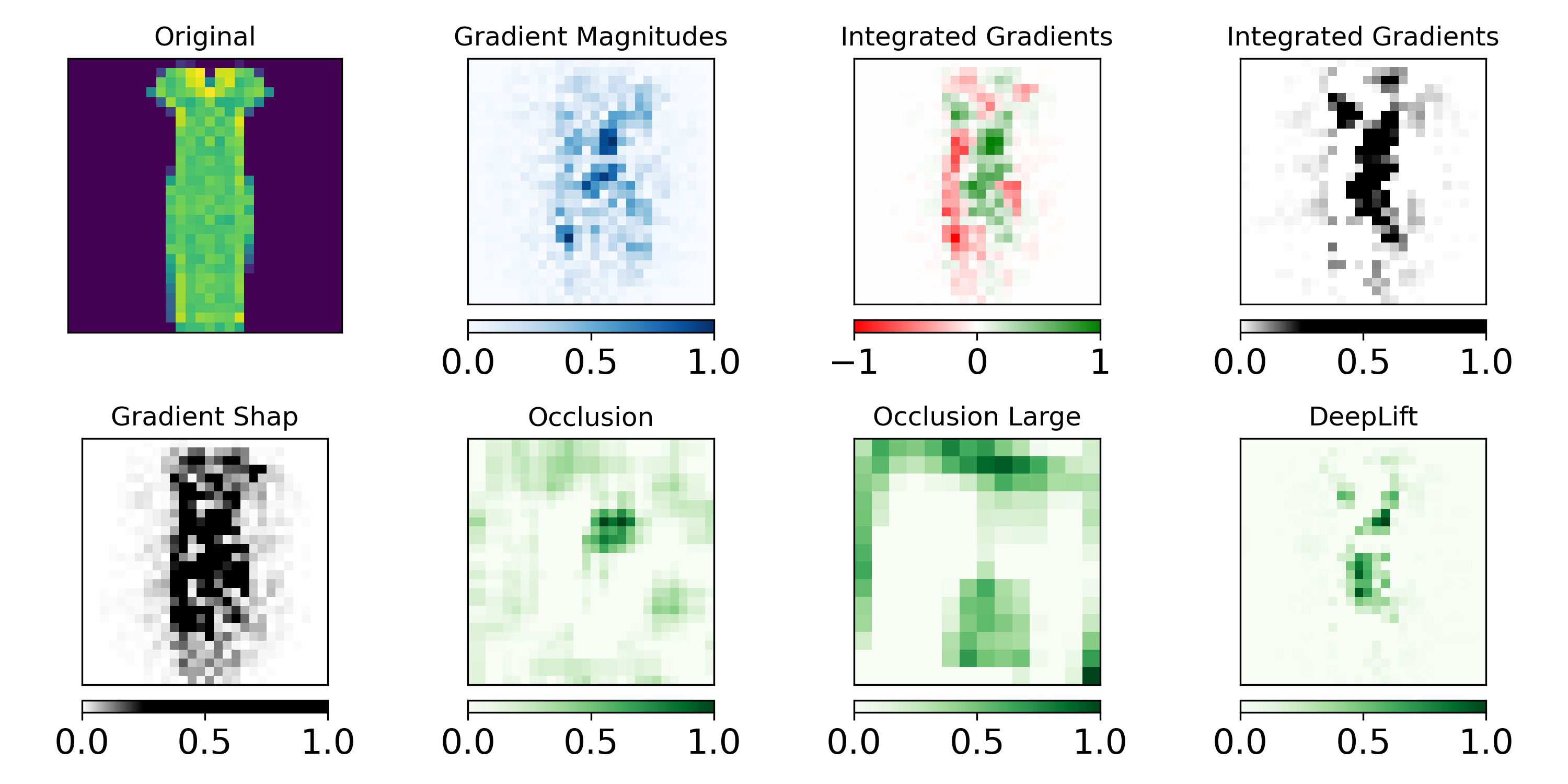}
\caption{\small fMNIST Dress, Model prediction 1. TRUST-LAPSE score = 0.21 (low).}
\label{attributions_fmnistdress_low}
\vskip -0.1in
\end{center}
\end{figure}

\begin{figure}[h]
\begin{center}
\includegraphics[width=\textwidth/2]{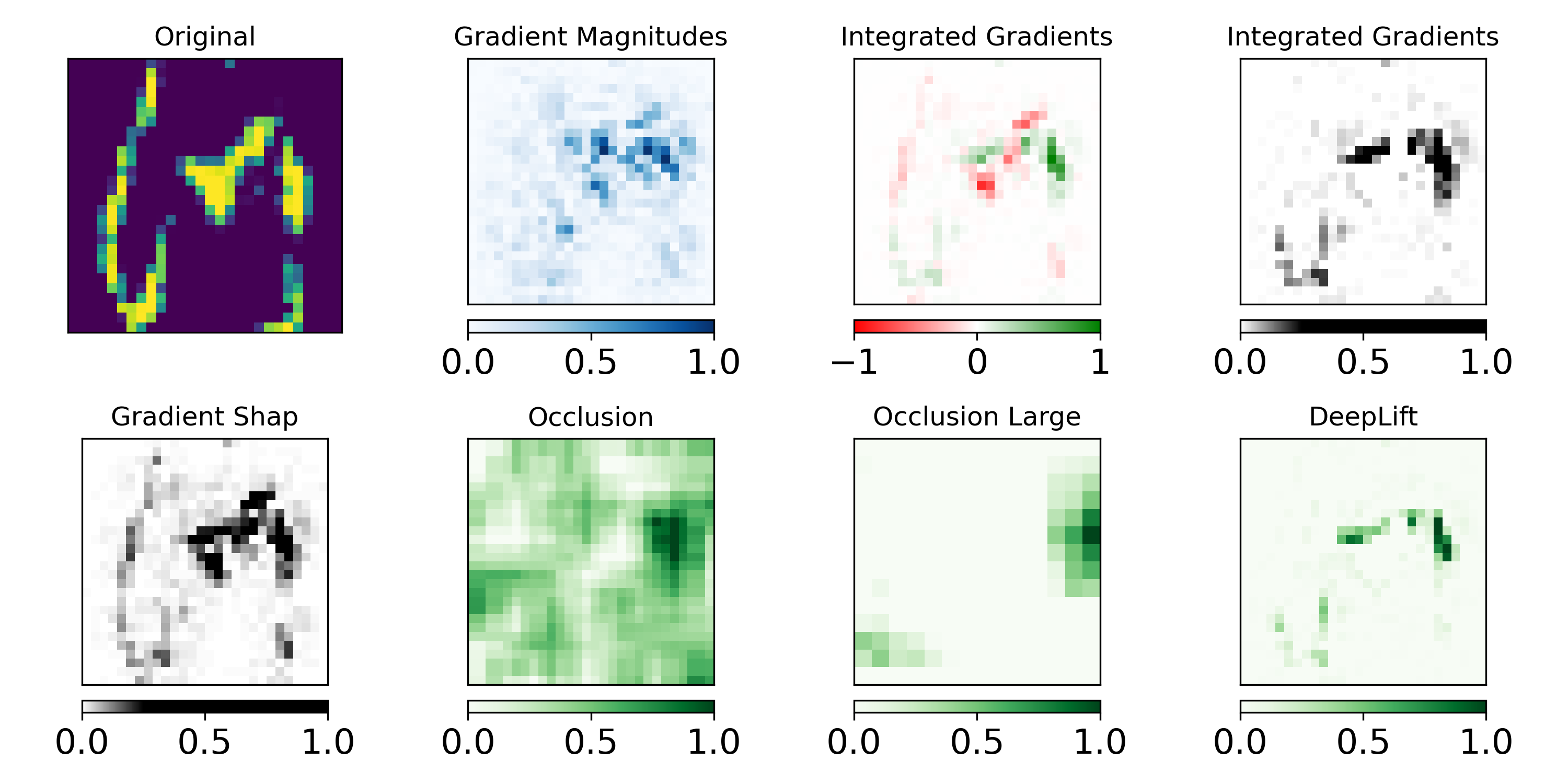}
\caption{\small kMNIST tsu, Model prediction 0. TRUST-LAPSE score = 0.06 (low).}
\label{attributions_kmnist_tsu1_low}
\vskip -0.1in
\end{center}
\end{figure}

\begin{figure}[h]
\begin{center}
\includegraphics[width=\textwidth/2]{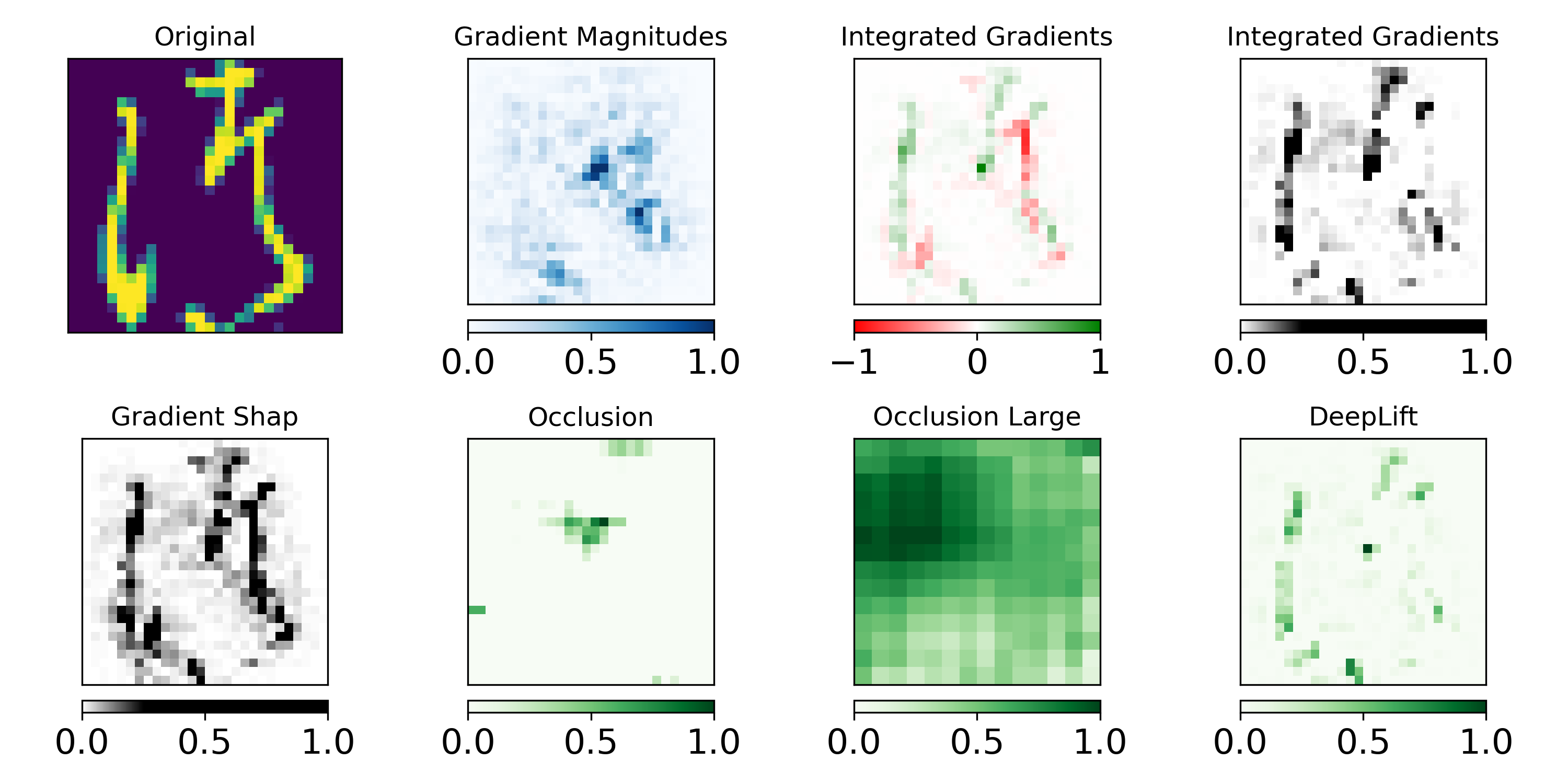}
\caption{\small kMNIST tsu, Model prediction 6. TRUST-LAPSE score = 0.10 (low).}
\label{attributions_kmnist_tsu2_low}
\vskip -0.1in
\end{center}
\end{figure}

\begin{figure}[h]
\begin{center}
\includegraphics[width=\textwidth/2]{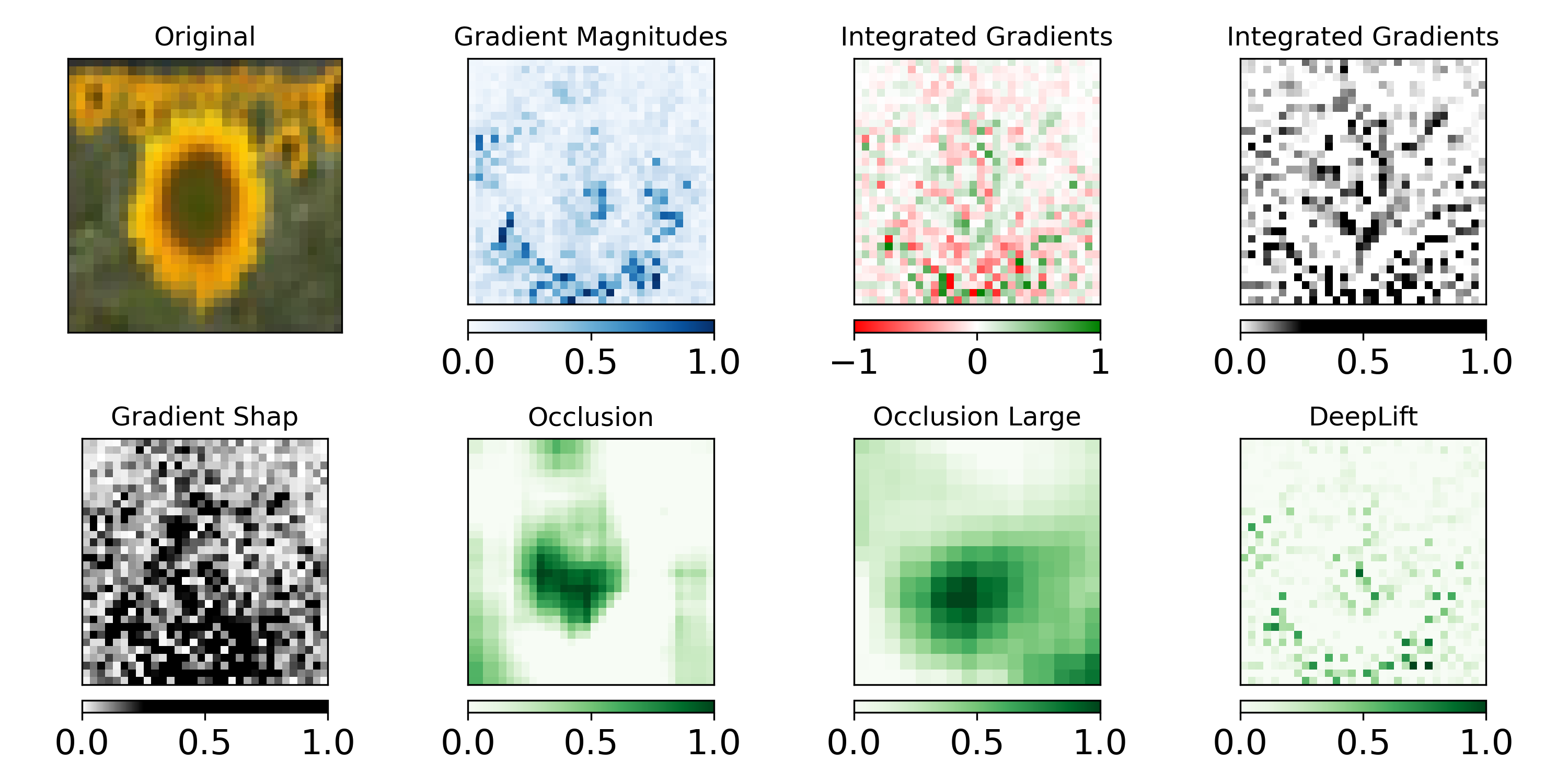}
\caption{\small CIFAR100 Sunflower, Model prediction Deer. TRUST-LAPSE score = 0.29 (low).}
\label{attributions_cifar100_sunflower_low}
\vskip -0.1in
\end{center}
\end{figure}

\begin{figure}[h]
\begin{center}
\includegraphics[width=\textwidth/2]{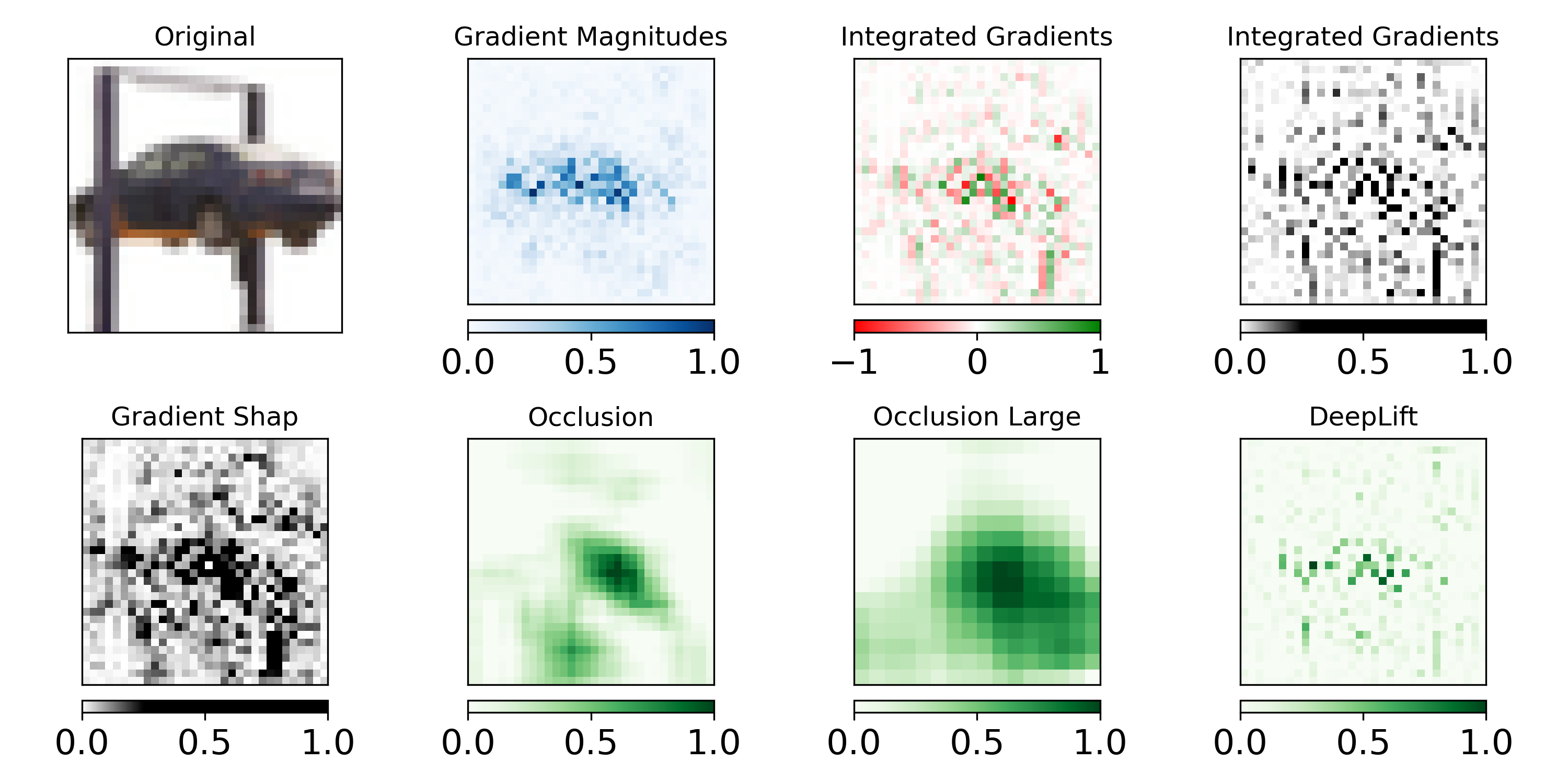}
\caption{\small CIFAR10 Automobile, Model prediction Horse. TRUST-LAPSE score = 0.45 (low).}
\label{attributions_cifar_automobile_low}
\vskip -0.1in
\end{center}
\end{figure}

\begin{figure}[h]
\begin{center}
\includegraphics[width=\textwidth/2]{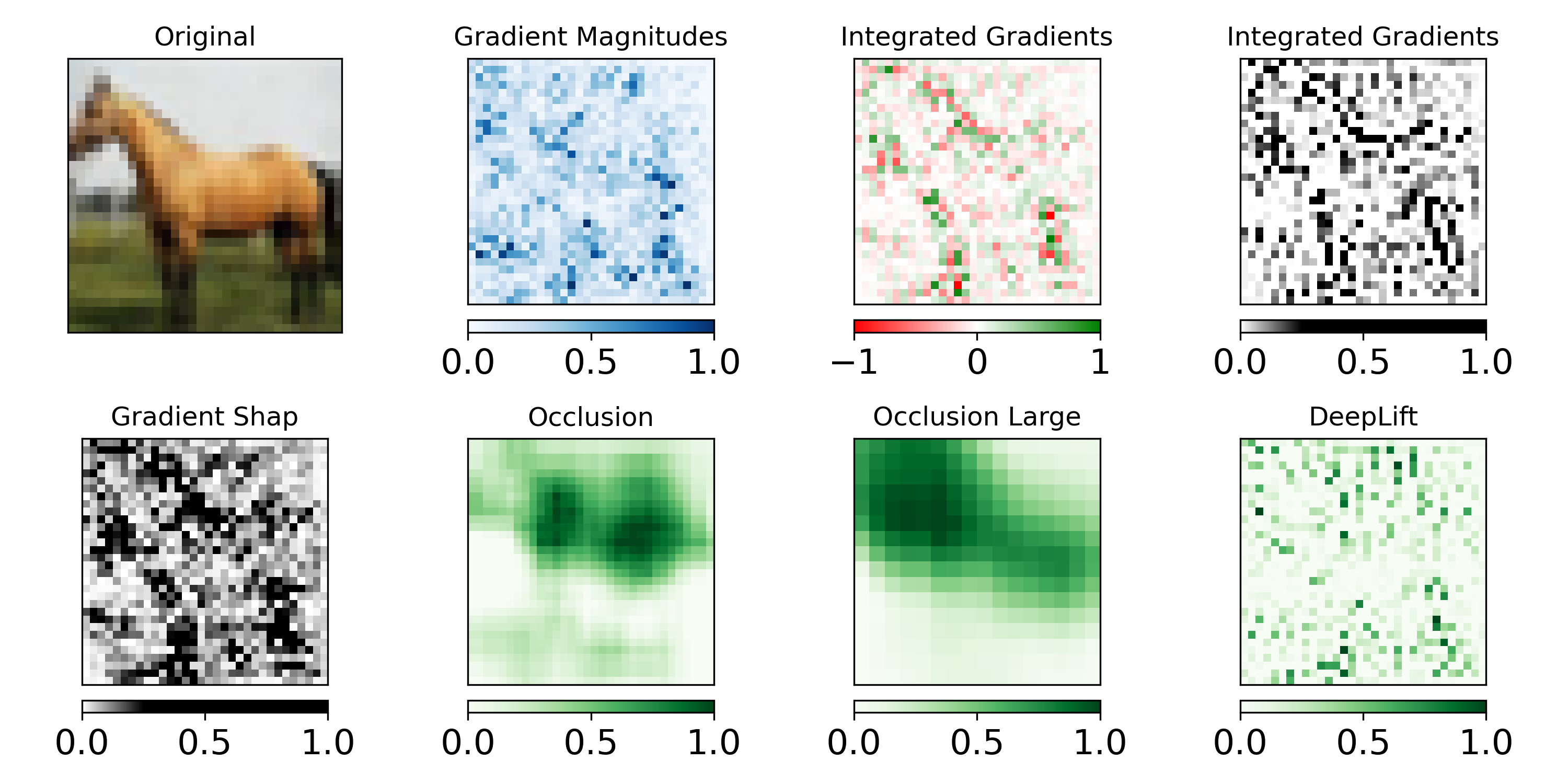}
\caption{\small CIFAR10 Horse, Model prediction Horse. TRUST-LAPSE score = 0.96 (high).}
\label{attributions_cifar_horse_high}
\vskip -0.1in
\end{center}
\end{figure}

Additional attribution examples for high and low TRUST-LAPSE Scores are given in Figs. \ref{attributions_mnist0_high}-\ref{attributions_cifar_horse_high}.\\

\subsubsection{\textbf{Latent-Space Visualizations}} 
\begin{figure*}[h]
\includegraphics[width=\textwidth]{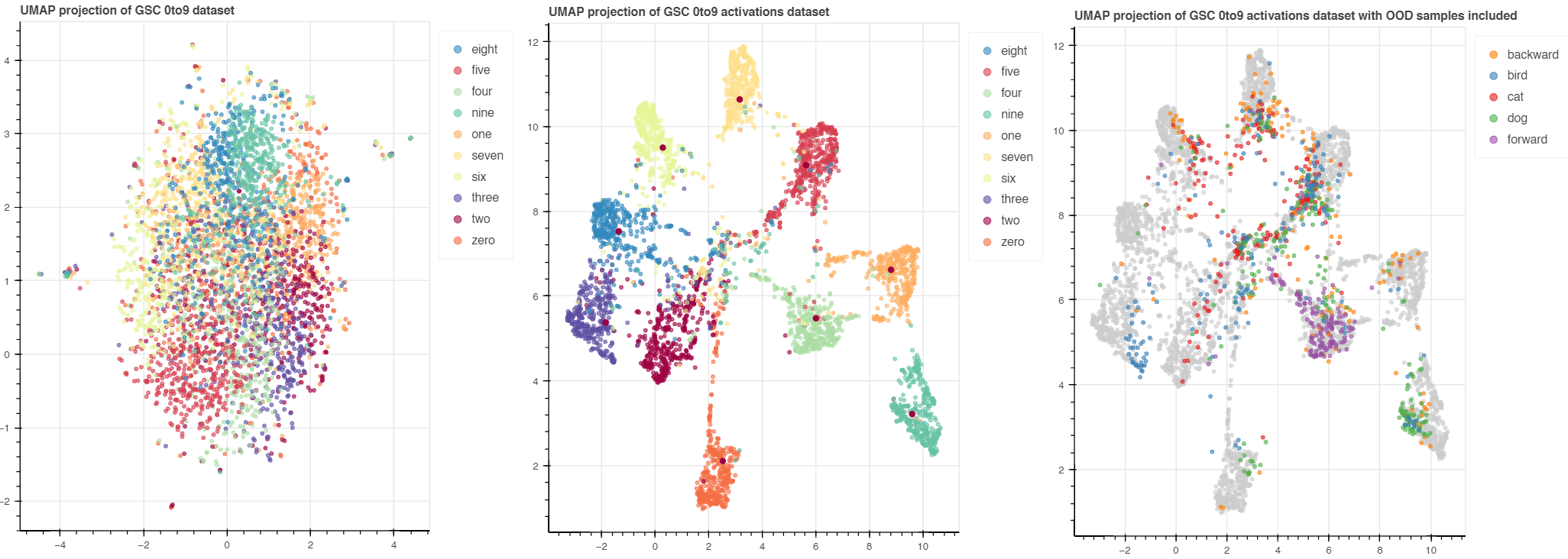}
\caption{\small   Latent-Space Visualization: UMAP embeddings of GSC audio data. (left) raw InD data (middle) representations from the trained encoder, i.e. a 2D projection of the latent-space (c) Semantic OOD samples added to the mix (only few classes shown).}
\label{gsc_umap}
\end{figure*}

\begin{figure*}
\includegraphics[width=\textwidth]{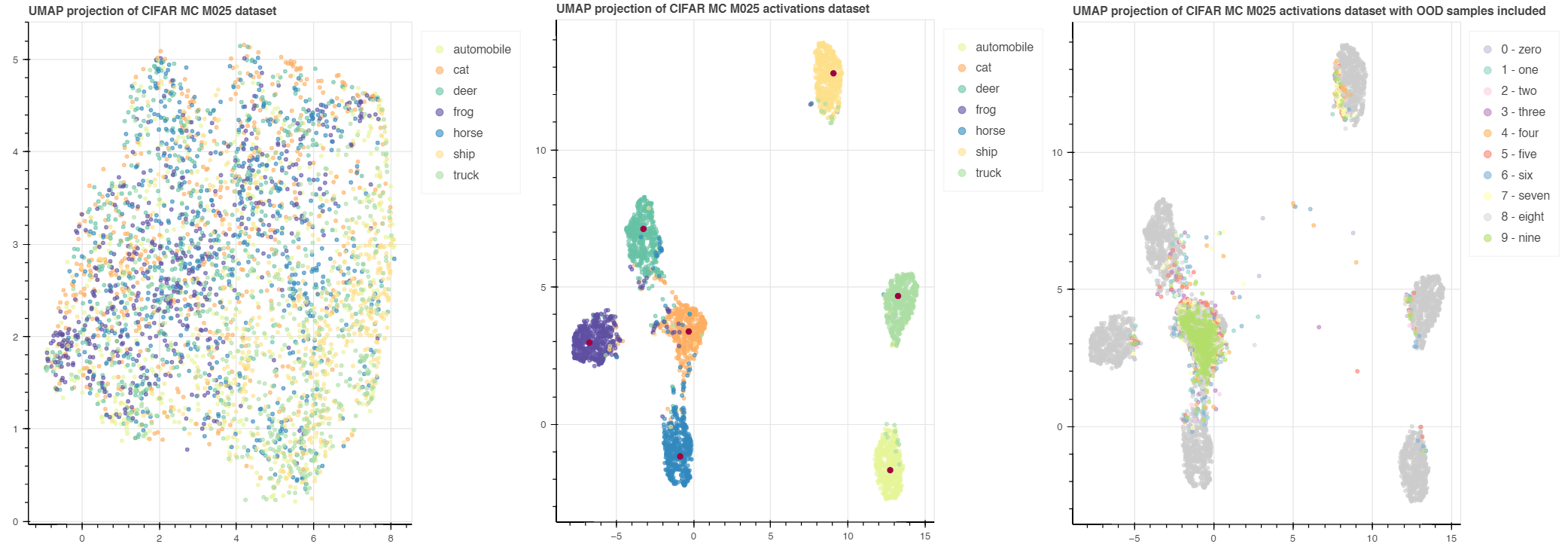}
\caption{\small    UMAP embeddings of CIFAR data (left) raw InD data (middle) representations from the trained encoder, i.e. a 2D projection of the latent-space (c) Semantic OOD samples added to the mix (only few classes are shown for visualization purposes).}
\label{cifar_umap}
\end{figure*}

\begin{figure*}
\includegraphics[width=\textwidth]{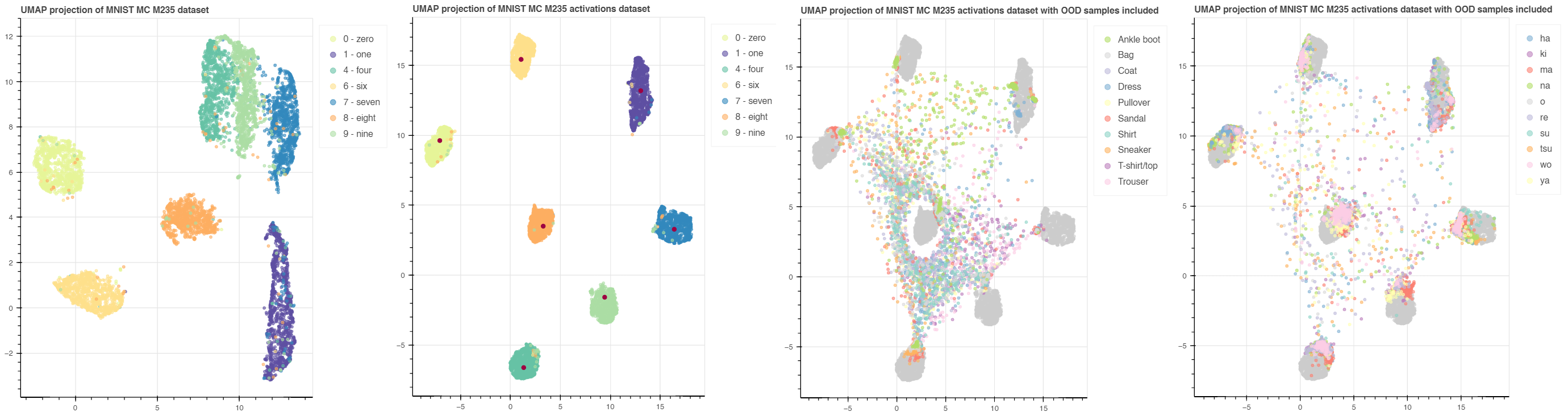}
\caption{\small    UMAP embeddings of MNIST data (left) raw InD data (middle) representations from the trained encoder, i.e. a 2D projection of the latent-space (c) Semantic OOD samples added to the mix (only few classes are shown for visualization purposes).}
\label{mnist_umap}
\end{figure*}

\begin{figure*}
\includegraphics[width=\textwidth]{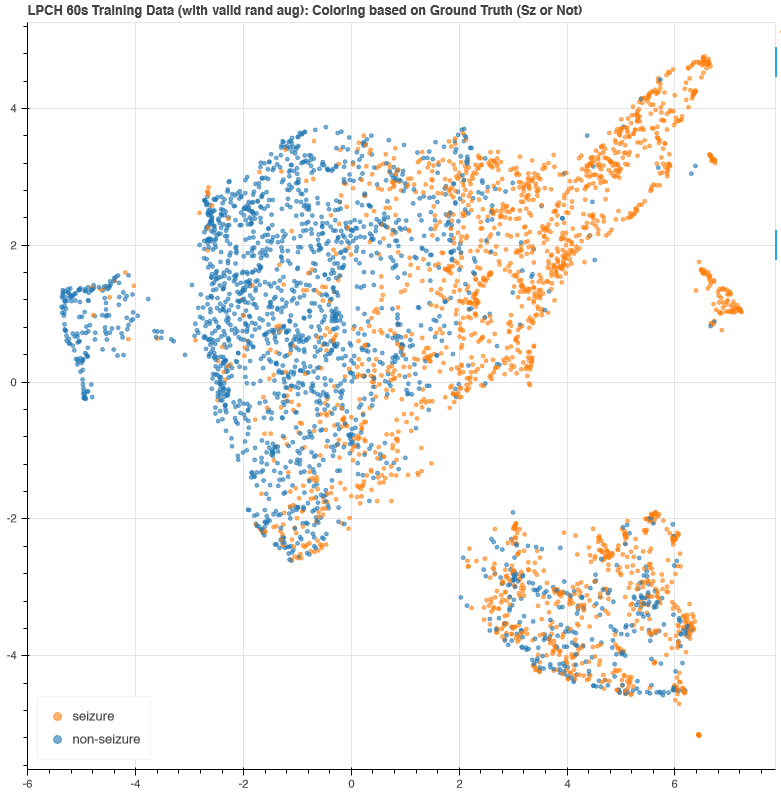}
\caption{\small    UMAP embeddings of EEG data: Representations from the trained encoder, i.e. a 2D projection of the latent-space are visualized here.}
\label{eeg_umap}
\end{figure*}

UMAP plots are shown in in Figs. \ref{gsc_umap}, \ref{cifar_umap}, \ref{mnist_umap}, \ref{eeg_umap}.\\

\subsubsection{\textbf{Distribution Plots}} 
\begin{figure*}
\includegraphics[width=\textwidth]{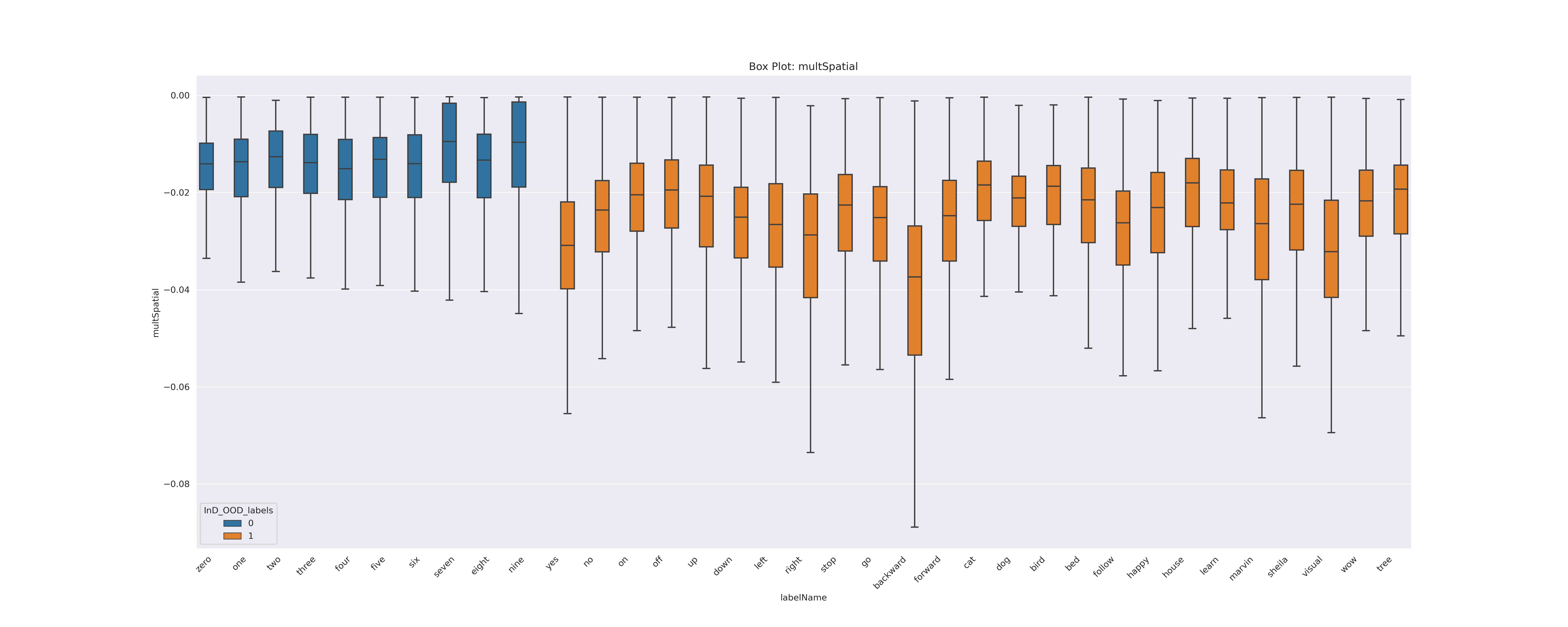}
\caption{\small     TRUST-LAPSE scores distribution on \edit{A}udio \edit{for the high capacity M5 encoder. The InD GSC 0-9 and FSDD 0-9 classes are merged together in blue and class-wise OOD scores from GSC-Words (some of them) are shown in orange.} InD and OOD classes are more separable.}
\label{gsc_score_distribution}
\end{figure*}

\begin{figure*}
\includegraphics[width=\textwidth]{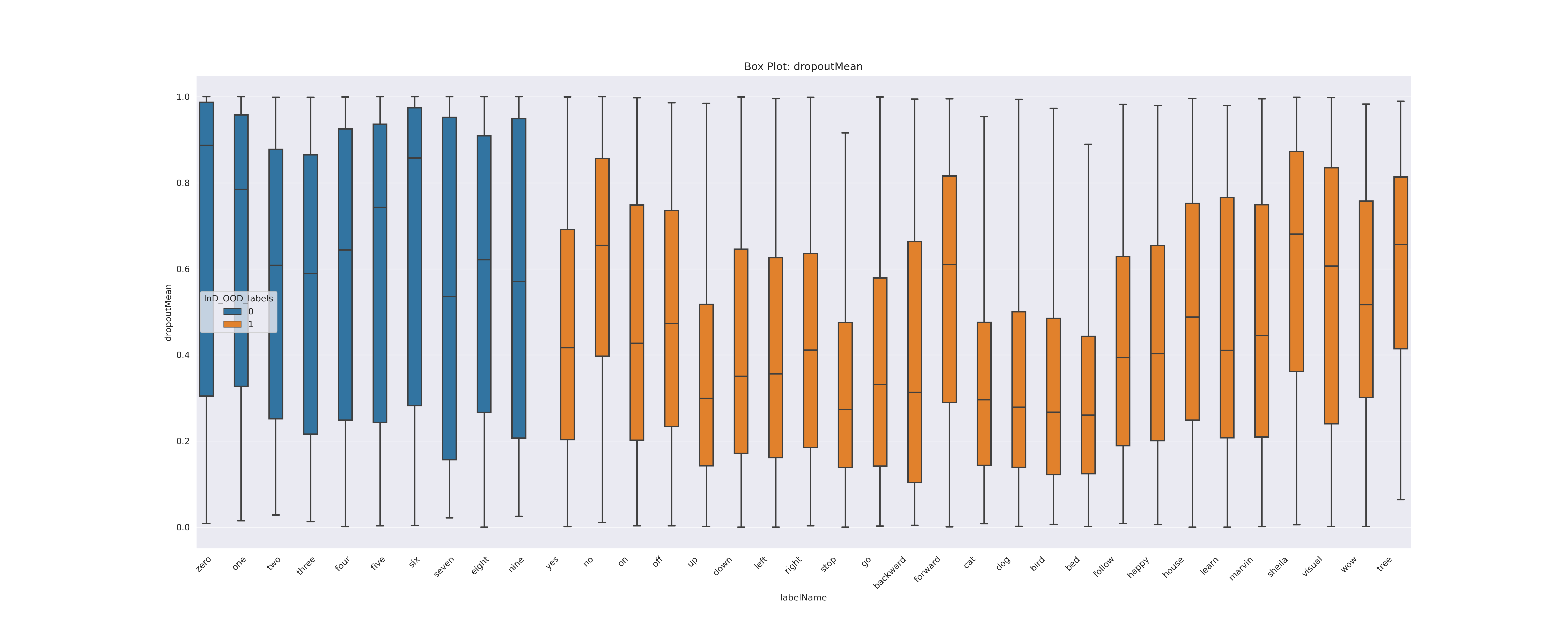}
\caption{\small    Test-time Dropout scores distribution on \edit{A}udio \edit{for the high capacity M5 encoder. The InD GSC 0-9 and FSDD 0-9 classes are merged together in blue and class-wise OOD scores from GSC-Words (some of them) are shown in orange.} InD and OOD classes are less separable.}
\label{test-time_dropout_score_distribution}
\end{figure*}

of latent-space mistrust scores from TRUST-LAPSE for GSC (audio) in Fig. \ref{gsc_score_distribution}. For comparison, test-time dropout score distributions are shown in Fig. \ref{test-time_dropout_score_distribution}. We see that InD and semantic OOD scores are much more separable for TRUST-LAPSE than Test-Time Dropout.
 
\end{document}